\pdfoutput=1

\documentclass[11pt]{article}

\usepackage{ACL2023}

\usepackage{times}
\usepackage{latexsym}

\usepackage[T1]{fontenc}

\usepackage[utf8]{inputenc}

\usepackage{microtype}

\usepackage{hyperref}
\usepackage{url}

\usepackage{graphicx}
\usepackage{xcolor}
\usepackage[most]{tcolorbox}
\usepackage{xspace}
\usepackage{booktabs}
\usepackage{algorithm}
\usepackage{algpseudocode}
\usepackage{amsfonts}
\usepackage{fdsymbol}
\usepackage{dsfont}
\usepackage{listings}
\usepackage{pifont}
\usepackage{tokcycle}
\usepackage{multirow}
\usepackage{enumitem}
\usepackage{sourcecodepro}
\usepackage{caption}
\usepackage{subcaption}
\usepackage{wrapfig}
\usepackage{setspace}
\usepackage{textcomp}

\definecolor{awesomeCyan}{RGB}{100,164,242}
\definecolor{awesomeMagenta}{RGB}{255,102,153}
\definecolor{awesomeGreen}{RGB}{89,206,86}
\definecolor{awesomeLightGray}{RGB}{240,240,240}
\definecolor{awesomeDarkGray}{RGB}{50,50,50}
\newtcbox{\inlinebox}[1][]{enhanced,
	box align=base,
	nobeforeafter,
	colback=awesomeLightGray,
	colframe=awesomeLightGray,
	size=small,
	left=0pt,
	right=0pt,
	boxsep=0.5pt,
	#1}
\newtcbox{\cyaninlinebox}[1][]{enhanced,
	box align=base,
	nobeforeafter,
	colback=awesomeCyan,
	colframe=awesomeCyan,
	size=small,
	left=0pt,
	right=0pt,
	boxsep=0.5pt,
	#1}
\newtcbox{\magentainlinebox}[1][]{enhanced,
	box align=base,
	nobeforeafter,
	colback=awesomeMagenta,
	colframe=awesomeMagenta,
	size=small,
	left=0pt,
	right=0pt,
	boxsep=0.5pt,
	#1}

\newcommand\mystrut{\rule[-0.3pt]{0pt}{0.7em}}
\newcommand{\token}[1]{\inlinebox{\color{awesomeDarkGray}{\mystrut\texttt{\footnotesize #1}}}\xspace}
\newcommand{\cyantoken}[1]{\cyaninlinebox{\color{white}{\mystrut\texttt{\footnotesize #1}}}\xspace}
\newcommand{\magentatoken}[1]{\magentainlinebox{\color{white}{\mystrut\texttt{\footnotesize #1}}}\xspace}
\newcommand{\GO}{\token{GO}}
\newcommand{\cGO}{\cyantoken{GO}}
\newcommand{\TAIL}{\token{TAIL}}
\newcommand{\cTAIL}{\cyantoken{TAIL}}
\newcommand{\STOP}{\token{STOP}}
\newcommand{\cSTOP}{\cyantoken{STOP}}
\newcommand{\THINK}{\token{THINK}}
\newcommand{\cTHINK}{\magentatoken{THINK}}
\newcommand{\PAD}{\token{PAD}}

\newcommand{\RMD}{\token{R}}
\newcommand{\TIMES}{\token{$\times$}}
\newcommand{\DIV}{\token{$\div$}}

\Characterdirective{\addcytoks{\token{#1}}}
\newcommand{\tokenize}[1]{\tokcyclexpress{#1}\the\cytoks}

\definecolor{codeGreen}{RGB}{71, 165, 69}
\definecolor{codeMagenta}{RGB}{224, 82, 122}
\definecolor{codegray}{rgb}{0.5,0.5,0.5}
\definecolor{codepurple}{rgb}{0.58,0,0.82}
\definecolor{backcolour}{rgb}{0.97,0.97,0.97}

\lstdefinestyle{mystyle}{
    backgroundcolor=\color{backcolour},   
    commentstyle=\color{codeGreen},
    keywordstyle=\color{codeMagenta},
    numberstyle=\scriptsize\color{codegray},
    stringstyle=\color{codeGreen},
    basicstyle=\ttfamily\scriptsize\setstretch{1.2},
    breakatwhitespace=false,         
    breaklines=true,                 
    captionpos=b,                    
    keepspaces=true,                 
    numbers=left,                    
    numbersep=10pt,                  
    showspaces=false,                
    showstringspaces=false,
    showtabs=false,                  
    tabsize=2,
    framesep=0pt,
    xleftmargin=35pt,
    xrightmargin=15pt,
    frame=tb,
    framexleftmargin=25pt,
    framexrightmargin=5pt,
    framexbottommargin=5pt,
    framextopmargin=5pt,
    upquote=true,
}

\usepackage{subfiles}

\title{Recursion of Thought: A Divide-and-Conquer Approach\\to Multi-Context Reasoning with Language Models}

\author{Soochan Lee \\
  Seoul National University \\
  \texttt{soochan.lee@vision.snu.ac.kr} \\\And
  Gunhee Kim \\
  Seoul National University \\
  SNU-LG AI Research Center \\
  \texttt{gunhee@snu.ac.kr} \\}

\begin{document}

\maketitle
\begin{abstract}
Generating intermediate steps, or Chain of Thought (CoT), is an effective way to significantly improve language models' (LM) multi-step reasoning capability.
However, the CoT lengths can grow rapidly with the problem complexity, easily exceeding the maximum context size.
Instead of increasing the context limit, which has already been heavily investigated, we explore an orthogonal direction: making LMs divide a problem into multiple contexts.
We propose a new inference framework, called Recursion of Thought (RoT), which introduces several special tokens that the models can output to trigger context-related operations.
Extensive experiments with multiple architectures including GPT-3 show that RoT dramatically improves LMs' inference capability to solve problems, whose solution consists of hundreds of thousands of tokens.
\end{abstract}

\section{Introduction}

Recently, LMs have become a prominent direction to solve reasoning.
Given a question sequence, the models are tasked to predict the following answer sequence.
One recent line of research for reasoning with LMs is \emph{chain of thought} (CoT) generation \citep{Nye2021ShowYW,Wei2022ChainOT,Kojima2022LargeLM}.
In CoT generation, complex reasoning problems are solved by generating intermediate reasoning steps, or chain of thought, before producing the final answer.
This allows the problem's complexity to be spread across multiple token generations, making each generation more straightforward given the previous tokens.

Although CoT dramatically increases reasoning accuracy, there is a critical issue that limits its utility: the effective context size of sequence models cannot grow unbounded.
Context refers to the set of input tokens that a model is conditioned on when generating output.
Practically, all sequence models have a limit on the maximum context length due to various reasons.
For instance, Transformers \citep{Vaswani2017AttentionIA} suffer from a quadratic computational cost on the context length, and RNNs \citep{Hochreiter1997LongSM} struggle with long-term dependency modeling.
Therefore, even the state-of-the-art LMs limit the maximum context length to a few thousand tokens.
However, complex real-world problems may take even millions of tokens of reasoning steps to reach the answer.

While there has been extensive research on Transformers with longer contexts \citep{Tay2020EfficientTA}, we explore an orthogonal direction: divide and conquer.
Our new model-agnostic inference framework \emph{Recursion of Thought} (RoT) lets an LM recursively create multiple contexts by producing special tokens.
Therefore, even if a problem's solution exceeds the maximum context size, the model can divide it into multiple short contexts.
We show the potential of RoT with our new synthetic benchmark consisting of eight arithmetic and algorithmic tasks.
One can easily adjust the difficulty of the tasks to produce problems with extremely long (100K+ tokens) reasoning steps.
Without any task-specific component, such as a calculator, the models with RoT can easily learn to solve extremely complex problems whose solutions consist of hundreds of thousands of tokens.
To the best of our knowledge, no previous work comes close to handling this scale of reasoning procedures.
Since RoT is an early exploration in this direction, it needs several improvements to be applied to more practical scenarios.
Nonetheless, the impressive experimental results suggest that the multi-context paradigm of RoT might play an important role in future LMs.
We also provide our PyTorch \citep{Paszke2019PyTorch} implementation that can \emph{fully} reproduce the experiments.\footnote{\url{https://github.com/soochan-lee/RoT}}

\section{Related Work}

Scratchpad \citep{Nye2021ShowYW} is one of the earliest approaches demonstrating that fine-tuning language models to produce CoT can largely improve reasoning accuracy.
In the paper, the authors also mention the confined context size as a major hurdle to scaling their method.
More recently, it has been found that sufficiently large pre-trained language models can be induced to produce CoT, by simply tuning the \emph{prompt} \citep{Wei2022ChainOT,Kojima2022LargeLM}.
Several concurrent works extend CoT prompting to decompose complex problems into smaller problems \citep{Dua2022SuccessivePF,Zhou2022LeasttoMostPE,Khot2022DecomposedPA}.
Although these works also share the principle of divide and conquer like RoT, they mostly focus on improving the reasoning accuracy of relatively small problems whose solutions usually can fit in a single context.
On the other hand, we focus on solving problems that the solutions are orders of magnitude longer than the context size.
More detailed description of related work can be found in Appendix \ref{sec:ext_related_work}.

\section{Recursion of Thought (RoT)}




\begin{figure*}[t]
  \centering
	\includegraphics[width=0.9\textwidth, trim={0 19cm 0 0}, clip]{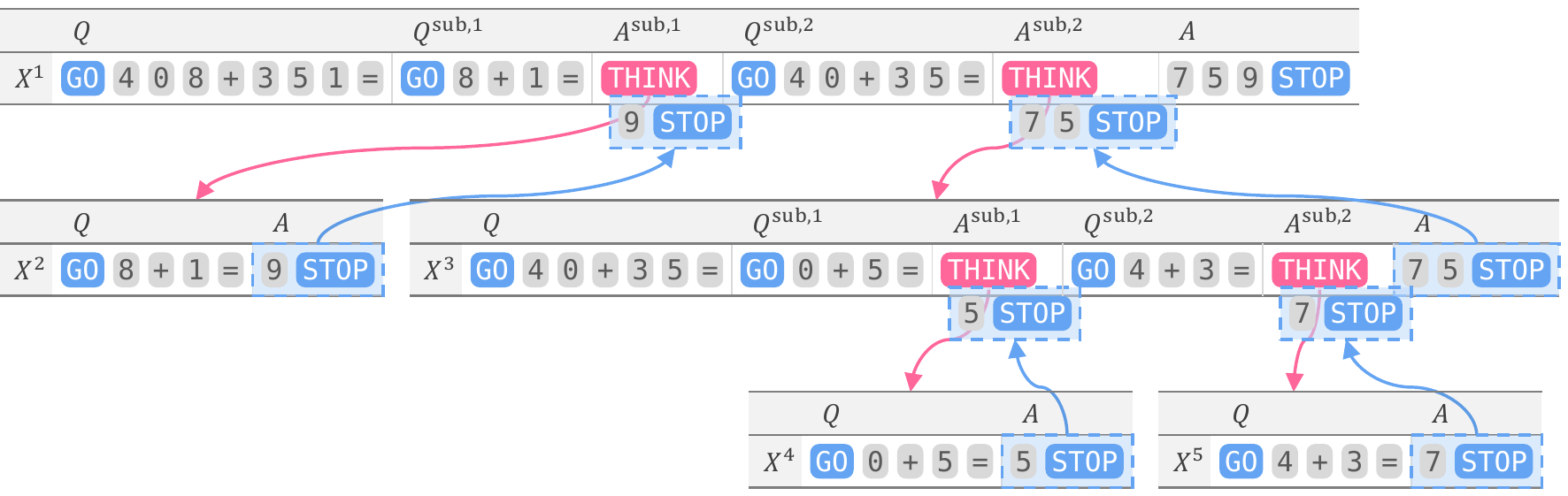}
	\caption{
		An example of the Recursion of Thought inference solving 408+351.
		Each table represents an inference context $X^k$ in order of creation.
		For each context, the model is given $Q$ and tasked to generate the rest.
		The model outputs the \THINK token when it needs to generate $A^{\mathrm{sub},*}$, the answer of a subproblem.
		The \THINK token triggers a recursive process and is later replaced by the answer returned from the process.
	}
	\label{fig:rot}
\end{figure*}

\subsection{Inference}
\label{sec:inference}

We start with how an RoT-trained LM performs at test time.
RoT is a model-agnostic framework, whose only requirement is that the model can infer $p(x_{i+1} | X_{1:i})$, the probability of the next token $x_{i+1}$ given a sequence $X_{1:i} = [x_1;...;x_i]$.
For recursive context control, we introduce the following special tokens: \GO, \STOP, and \THINK.
\GO and \STOP respectively mark the start and end of a problem sequence.
They can be nested inside another \GO-\STOP pair to indicate a subproblem.
\THINK initiates a recursion procedure.
RoT teaches a model how to use these tokens so that it can perform divide-and-conquer problem-solving.
We formulate each inference context of a QA problem, denoted $X$, as the following concatenation:
\begin{equation}
	\label{eq:context}
	\begin{split}
		X = [Q; Q^{\mathrm{sub},1}; A^{\mathrm{sub},1}; \ldots; Q^{\mathrm{sub},N}; A^{\mathrm{sub},N}; A]
	\end{split}
\end{equation}
where $Q$ and $A$ are the main question and answer sequence, and $Q^{\mathrm{sub},*}$ and $A^{\mathrm{sub},*}$ are those of the \emph{top-level} subproblems.
Although a subproblem can have smaller, lower-level subproblems recursively, only the top-level subproblems remain in an RoT context.
During inference, a model is given $Q$ and tasked to generate the rest.
Questions ($Q$ and  $Q^{\mathrm{sub},*}$) start with a \GO token, and answers ($A$ and  $A^{\mathrm{sub},*}$) end with a \STOP token.
In the base cases, contexts do not have $(Q^{\mathrm{sub},*}, A^{\mathrm{sub},*})$ pairs.

Figure \ref{fig:rot} presents an example of solving $408+351$ for better understanding.
The pseudocode and more detailed illustrations can be found in Appendix \ref{sec:rot_alg} and \ref{sec:illustrated_rot}.
RoT starts by initializing the context $X$ with the original question $Q$ (i.e., \GO\tokenize{408+351=} in Figure \ref{fig:rot}).
Then, similar to CoT, the model solves multiple subproblems (generating $Q^{\mathrm{sub},*}$ and $A^{\mathrm{sub},*}$) before producing the final answer.
However, there is a key difference: instead of producing a sub-answer directly, the model outputs the \THINK token.
This special token triggers a recursive process that separates the sub-question in a new context.
If the new context is a base case (i.e., $X^2$, $X^4$, and $X^5$), the answer is produced directly.
Otherwise, the model recursively solves more subproblems.
If enough subproblems are solved, the model generates the final answer ending with a \STOP.
Once an answer is returned to the previous context, it replaces the \THINK token, and the generation continues.

For tail recursion, where the last subquestion's answer becomes the final answer, we additionally introduce the \TAIL token.
If \TAIL is used in the place of a \GO token in the last subquestion $Q^{\mathrm{sub},N}$, its answer $A^{\mathrm{sub},N}$ is treated as the final answer $A$.
Tail recursion is crucial since it enables an indefinitely long chain of recursion without overflowing the call stack.

\paragraph{The generality of RoT.}
Recursion, the core of RoT, is an incredibly general concept that serves as a fundamental building block for functional programming languages.
Any non-recursive procedure can be converted to a recursive form via continuation-passing style.
Therefore, there is no theoretical limit on the range of problems that RoT can handle.

\subsection{Training}
\label{sec:training}

Currently, we train RoT in a supervised manner, using ground truth (GT) intermediate steps that include when to output the special tokens.
The GTs are constructed following the standard procedures developed for humans.
For example, the procedures for arithmetic problems are borrowed from elementary school math.
More details can be found in Appendix \ref{sec:data_gen}.
We leave training RoT with less supervision as a future work.

Each training example is constructed as a pair of a ground truth context sequence $X$ and the corresponding target sequence $Y$.
An example and the pseudocode for creating a target sequence are presented in Figure \ref{fig:target} and Algorithm \ref{alg:context_conversion} in Appendix \ref{sec:target_seq}.
Overall, $Y$ is a copy of $X$ except for the parts corresponding to $Q$ and $A^{\mathrm{sub},*}$. 
Since the question $Q$ is always given in a context, $Q$ is replaced by special \PAD tokens, which are excluded from the loss function.
Each subproblem's answer $A^{\mathrm{sub},n}$ is replaced by a \THINK token followed by several {\PAD}s that fill in the rest to make sure $|X|=|Y|$.
This way, the model is trained to output \THINK instead of the first token of $A^{\mathrm{sub},n}$.
Since the whole $A^{\mathrm{sub},n}$ will be returned from the recursive process and replace the \THINK during inference, we do not need a training signal for the rest of $A^{\mathrm{sub},n}$.

Given a pair $(X, Y)$, the training objective is defined as follows:
\begin{equation}
	\mathcal L = -\sum_i I[y_{i+1} \neq \PAD] \log p(y_{i+1} | X_{1:i})
\end{equation}
where $I$ is the indicator function that excludes {\PAD}s from training.
Its form is almost identical to the standard LM objective: 
$\mathcal L_{\mathrm{LM}} = -\sum_i \log p(x_{i+1} | X_{1:i})$,
which is to predict the next token given previous tokens.
Therefore, any sequence model is trained in the standard way, i.e., end-to-end via stochastic gradient descent.

\section{Experiments}

\begin{figure*}[t]
  \includegraphics[width=\textwidth, trim={0 5.7cm 0 0}, clip]{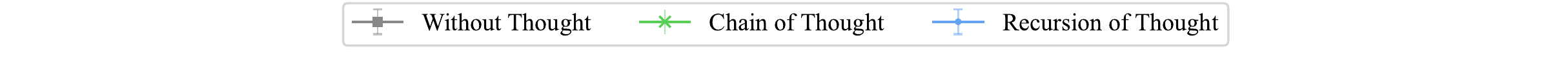}
	\begin{subfigure}[b]{0.35\textwidth}
    \centering
		\includegraphics[height=60mm]{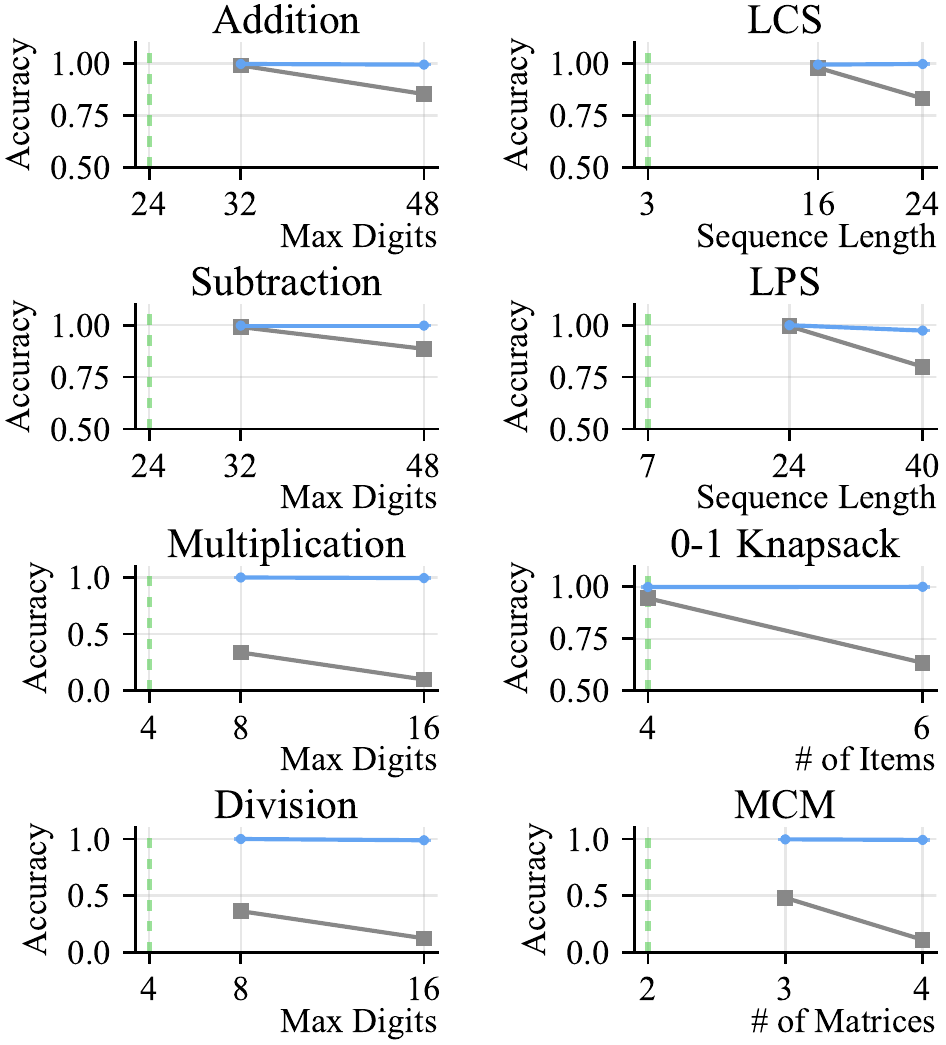}
		\caption{GPT-3}
		\label{fig:exp:gpt3}
	\end{subfigure}
  \hspace{2mm}
	\begin{subfigure}[b]{0.4\textwidth}
    \centering
		\includegraphics[height=60mm]{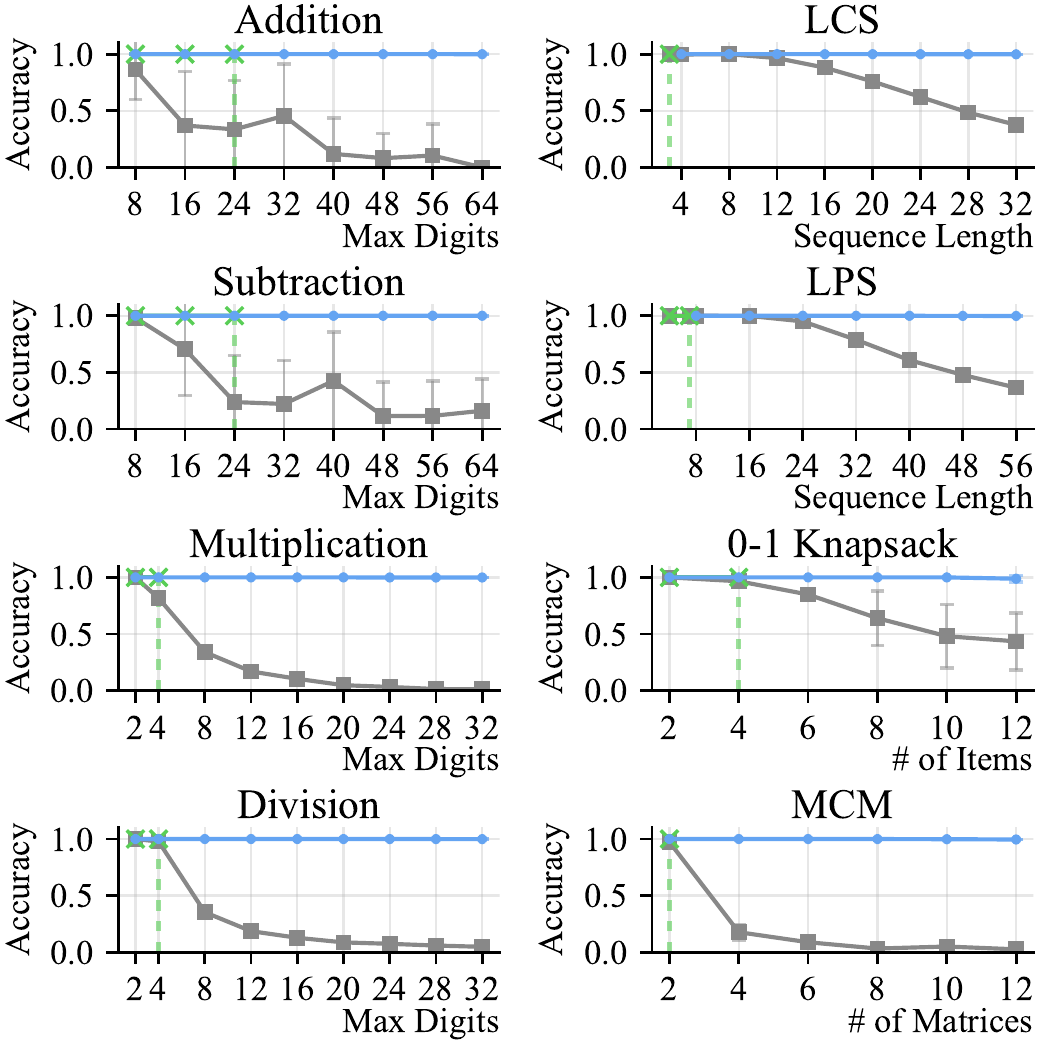}
		\caption{Tiny Transformer}
		\label{fig:exp:transformer}
	\end{subfigure}
  \hspace{2mm}
	\begin{subfigure}[b]{0.2\textwidth}
    \centering
		\includegraphics[height=60mm]{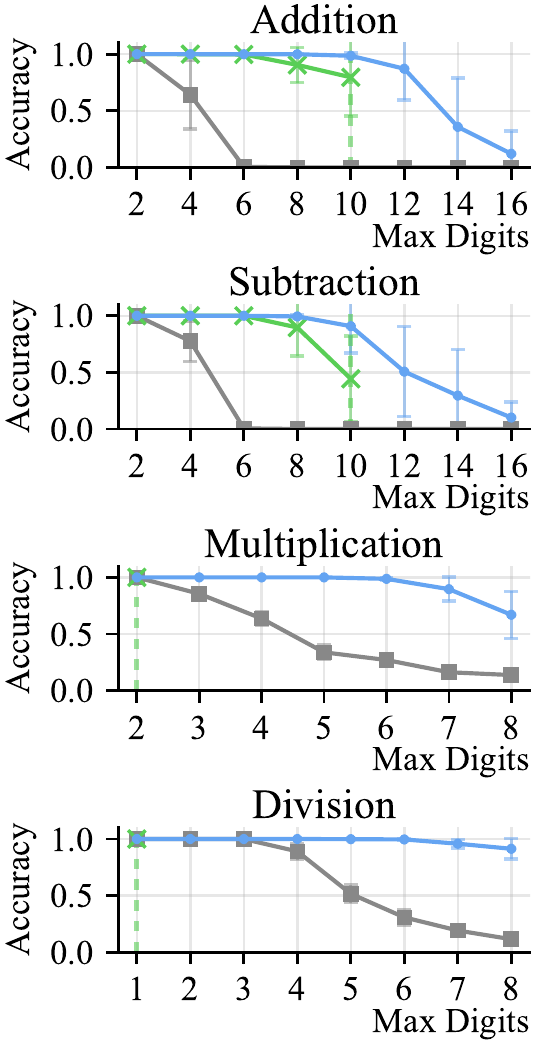}
		\caption{Tiny LSTM}
		\label{fig:exp:lstm}
	\end{subfigure}
	\caption{
		Comparison of the thought types.
		In each graph, the x-axis is the problem difficulty, while the y-axis is the reasoning accuracy.
		Each point represents an independent experiment.
		The green vertical lines indicate the maximum problem difficulty that CoT can handle without exceeding the maximum context size.
	}
	\label{fig:experiments}
\end{figure*}

\subsection{Baselines}
We compare RoT with two baselines.
The first one is to output an answer directly from a question, which we call \emph{Without Thought} (WT).
The other one is to generate all the intermediate steps before the answer without recursion \citep{Nye2021ShowYW}, which we refer to as \emph{Chain of Thought} (CoT; not to be confused with the CoT \emph{prompting} \citep{Wei2022ChainOT}). 
We construct the ground truths for CoTs by unraveling the same recursive process which we design for RoT, into a single context sequence (see Appendix \ref{sec:cot_example} for examples).
Therefore, the number of tokens to generate while solving a problem is the same for both CoT and RoT (if we do not count the \THINK tokens).
However, the sizes of the individual contexts of CoT are far longer than those of RoT due to the recursively nested subproblems, limiting the range of solvable problems.
Refer to Appendix \ref{sec:ctx_length} for a more detailed analysis of the context sizes.
For a fair comparison, we \emph{train} these baselines and do not use any prompting technique.
When evaluating, we consider a problem to be correctly solved only if all the intermediate steps and the answer are correct.

\subsection{The Reasoning Problems}
\label{sec:probs}

To evaluate the reasoning capabilities, we test four basic arithmetic tasks and four algorithmic tasks: addition, subtraction, multiplication, division, longest common subsequence, longest palindromic subsequence, 0-1 knapsack, and matrix chain multiplication.
The details can be found in Appendix \ref{sec:prob_spec}.
We choose these tasks because we can easily increase the problem difficulty while being able to get ground truth solutions.
Therefore, we can test problems whose solution contains hundreds of thousands of tokens.
All problems are formulated in pure sequence modeling, without any external programs (e.g., calculator) involved.

\subsection{Experiments with GPT-3}

Using the OpenAI API, we fine-tune GPT-3 for each reasoning task in \S\ref{sec:probs} for 10K steps with a batch size of 256.
The results are presented in Figure \ref{fig:exp:gpt3}, and the technical details are described in Appendix \ref{sec:gpt-3}.
Each point in the graphs represents one experiment at a certain problem difficulty.
We report the accuracy on a test set of 1K unique problems randomly sampled as explained in Appendix \ref{sec:prob_spec}.
To the best of our knowledge, the problems at this scale (e.g., 48-digit addition/subtraction and 16-digit multiplication/division) have never been solved by any LM without the help of external programs.
For reference, Minerva \citep{Lewkowycz2022SolvingQR} achieves around 80\% accuracy on 10-digit addition and 20\% on 18-digit addition.

\paragraph{Results.}
Even WT fine-tuning cannot make GPT-3 deal with such a level of complexity, while CoT is not applicable due to the context limit of 2048.
The green dotted lines mark the maximum difficulty that can be handled by CoT under the context limit.
On the other hand, RoT enables the GPT-3 to achieve near-perfect scores in every experiment.
As presented in Appendix \ref{sec:ctx_length}, solving each problem requires up to tens of thousands of tokens.
Without any architectural change, RoT makes GPT-3 handle these extremely complex problems.

\subsection{Experiments with Tiny Language Models}
Recent research on reasoning has been mostly focused on extremely large pre-trained LMs.
In this section, we show an interesting result that RoT can make even tiny models, without any pre-training, perform convoluted reasoning procedures.
We test the two basic sequence model architectures: a Transformer \cite{Vaswani2017AttentionIA} with 536K parameters and an LSTM \cite{Hochreiter1997LongSM} with 272K parameters.
These models are more than a million times smaller than the recent 540B-parameter PaLM \citep{Chowdhery2022PaLMSL}.
The context limit is set to 2048 for the Transformer and 512 for the LSTM.

By virtue of their small sizes, we conduct far more extensive experiments than GPT-3, which are presented in Figure \ref{fig:exp:transformer} and Figure \ref{fig:exp:lstm}.
For each experiment, we train a randomly initialized model and evaluate it on a test set of 30K unique problems.
We repeat each experiment eight times and report the average and standard deviation of the accuracies.
With the tiny Transformer, we experiment to the extent that even humans would find daunting.
For example, we test addition/subtraction up to 64 digits and multiplication/division up to 32 digits.
Note that a 32-digit number cannot even fit into the 64-bit integer datatype.

Throughout the experiments, we observe consistent patterns:
(i) WT's accuracy drops most quickly as the problem difficulty increases,
(ii) CoT achieves near-perfect accuracy, but it can only be applied to simple problems due to the context limit,
(iii) RoT achieves near-perfect accuracy and can be scaled up to extremely complex problems.
Despite the small sizes, RoT makes the Transformers master all types of extremely complex problems.
We do not test more difficult problems mainly because the evaluation becomes too costly, not because RoT is incapable of learning them.

\section{Conclusion}

We explored the novel idea of making LMs produce special tokens to create multiple contexts.
Following the principle of divide and conquer, LMs with RoT can solve extremely complex problems that have never been handled by any LM.
We believe the core idea of utilizing multiple contexts has a great potential and can play an essential role in future language models.

\section*{Limitations}

Although RoT remarkably improves LMs' reasoning capability, we currently rely on supervised training to teach RoT.
To apply RoT to a wider range of tasks, it would be crucial to reduce the expensive supervision.
Parallel to our work, \citet{Khot2022DecomposedPA} use prompting techniques to induce LMs to decompose problems.
However, prompting has other drawbacks.
First, lengthy prompts should be added for each inference, causing additional computational overhead.
And more critically, it is hard to guarantee high accuracy.
To achieve reasonable accuracy in the tasks in our experiments, each subproblem should be solved at extremely high accuracy (e.g., > 99.9\%) since each problem may contain hundreds or thousands of subproblems.
We have tested several prompting techniques with GPT-3, but could not get satisfactory accuracy.
Therefore, we conclude that solely relying on prompting cannot be a solution to this problem.
As one possible approach, we may combine RoT with the RL-based methodologies that are developed for reducing supervision of Neural Programmer-Interpreters \citep{Li2017NeuralPL, Fox2018ParametrizedHP, Pierrot2019LearningCN}.

Another limitation of this work is that the experiments are performed on somewhat synthetic tasks.
Since our goal is to enable LMs to solve reasoning problems whose intermediate steps are orders of magnitude longer than the context limit, we need a dataset with such complex problems.
However, no currently available dataset meets this requirement.
For example, the Long-Range Arena benchmark \citep{Tay2020LongRA} has at most 16K-token sequences and focuses on problems with long inputs and short outputs.
On the other hand, we tackle problems that require generating 100K+ tokens to solve.
Gathering natural language data at this scale is extremely challenging and costly.
Therefore, we currently resort to arithmetic and algorithmic problems since it is easy to scale them up and generate ground-truth solutions.
In the future, we hope to see new datasets and benchmarks that cover natural language reasoning at this scale.

Interestingly, RoT cannot facilitate length generalization, e.g., training on 8-digit multiplication with RoT cannot make a model generalize to 16-digit multiplication.
We believe this problem is rooted in a more fundamental limitation of the Transformer architecture \citep{Hahn2020TheoreticalLO}, orthogonal to RoT.
Fortunately, since RoT is a model-agnostic framework, we would be able to apply RoT to more advanced architectures to come in the future, which might be capable of length generalization.

\section*{Ethics Statement}

Since the problem types in our experiments are pure arithmetic or algorithmic tasks, we do not find any ethical concerns directly related to our work.
If RoT is applied to more general problems, the training data should meet ethical standards to ensure the non-toxic behavior of the model.

\section*{Acknowledgements}
We thank Jaekyeom Kim, Hyunwoo Kim, and Dongjoo Kim for their thoughtful discussions.
This work is partly supported by LG AI Research, 
the Institute of Information \& Communications Technology Planning \& Evaluation (IITP) grant funded by the Korea government (MSIT) (No.2019-0-01082, SW StarLab; No.2022-0-00156, Fundamental research on continual meta-learning for quality enhancement of casual videos and their 3D metaverse transformation), 
and the National Research Foundation of Korea (NRF) grant funded by the Korea government (MSIT) (No.2023R1A2C2005573).


\bibliography{anthology,rot}
\bibliographystyle{acl_natbib}

\newpage
\onecolumn
\appendix

\section{Extended Related Work}
\label{sec:ext_related_work}


\paragraph{Chain of Thought.}
Scratchpad \citep{Nye2021ShowYW} fine-tunes LMs to generate CoT before the final answer.
It demonstrates its effectiveness in 8-digit addition, polynomial evaluation, and Python program execution.
Instead of fine-tuning, it is found that we can elicit large pre-trained LMs to produce CoT with appropriate prompting.
For example, CoT prompting \citep{Wei2022ChainOT} adds several QA exemplars with CoT before the main question, encouraging the model to generate final answers in a similar manner.
Compared to the few-shot CoT prompting of \citet{Wei2022ChainOT}, \citet{Kojima2022LargeLM}'s zero-shot CoT prompting is even simpler;
after a question, they start the answer with ``Let's think step by step,'' and then let the model finish the rest.
Minerva \citep{Lewkowycz2022SolvingQR} utilizes these prompting techniques with a specially curated scientific pre-training dataset to achieve remarkable results on various reasoning benchmarks.

\paragraph{Prompting Language Models to Divide and Conquer Reasoning Problems.}
Based on CoT prompting \citep{Wei2022ChainOT}, several concurrent works demonstrate that decomposing problems into smaller subproblems can effectively improve reasoning accuracy.
Successive prompting \citep{Dua2022SuccessivePF} induces a model to alternate between generating a question and answering the question until the final answer is produced.
Similarly, least-to-most prompting \citep{Zhou2022LeasttoMostPE} makes a model start from the easiest subproblem and progressively solve more complex ones on top of the previous results.
Decomposed prompting \citep{Khot2022DecomposedPA} is a modular approach that the subproblems are solved by different modules depending on the problem type.
It also supports recursive decomposition.
These works are all closely related to our work.
Our work is unique in that we deal with far more complex problems that consist of thousands of subproblems.
In this case, the individual subproblems should be solved with almost perfect accuracy, or the overall accuracy drops significantly.
We empirically find that such a level of accuracy is hard to achieve by simply prompting a pre-trained LM.

\paragraph{Neural Programmer-Interpreter (NPI).}
Unlike language models, NPI \citep{Reed2016NeuralP} interacts with its environment through a series of program execution.
It consists of an LSTM core, an encoder for each domain, and a memory of program embeddings.
At every time step, the LSTM core takes a program embedding, arguments, and an observation of its environment to produce the next program embedding and corresponding arguments.
\citet{Cai2017MakingNP} combine NPI with recursion and show that recursion plays a critical role in generalization.
Since NPI requires full execution traces for training, there are multiple works to relax this requirement using reinforcement learning \citep{Li2017NeuralPL, Fox2018ParametrizedHP, Pierrot2019LearningCN}.


\section{RoT Inference Algorithm}
\label{sec:rot_alg}

\begin{algorithm}[H]
  \caption{Recursion of Thought Inference}
  \label{alg:rot}
  \begin{algorithmic}[1]
    \Require{A sequence model $\mathcal M$ trained for Recursion of Thought, a question sequence $Q$}
    \Function{RoT}{$\mathcal M$, $Q$} \label{alg:rot:start}
    \State $X \gets Q$ \Comment{Initialize context with $Q$}
    \State $i_\mathrm{ans} \gets |X|+1$ \Comment{Start of answer}
    \State $t \gets false$  \Comment{Tail recursion}
    \While{True}
    \State $x \gets \mathcal M(X)$ \Comment{Generate next token}
    \State $X \gets $ $[X; x]$
    \If{$x = \STOP$}
    \State \Return $X_{i_\mathrm{ans}:|X|}$ \label{alg:rot:return}
    \ElsIf{$x = \GO$}
    \State $i_{\mathrm{go}} \gets |X|$ \Comment{Mark last \GO} \label{alg:rot:go}
    \ElsIf{$x = \TAIL$}
    \State $i_{\mathrm{go}} \gets |X|$
    \State $t \gets true$ \label{alg:rot:mark_tail} \Comment{Mark tail recursion}
    \ElsIf{$x = \THINK$}
    \State $Q^\mathrm{sub} \gets X_{i_{\mathrm{go}}: |X|-1}$ \label{alg:rot:q_sub}
    \State $A^\mathrm{sub} \gets \Call{RoT}{\mathcal M, Q^\mathrm{sub}}$ \label{alg:rot:call}
    \If{$t$}
    \State \Return $A^\mathrm{sub}$ \label{alg:rot:return_tail}
    \EndIf
    \State $X \gets [X_{1:|X|-1}; A^\mathrm{sub}]$ \label{alg:rot:replace}
    \State \Comment{Replace \THINK with $A^{\mathrm{sub}}$}
    \State $i_{\mathrm{ans}} \gets |X|+1$ \label{alg:rot:i_ans}
    \EndIf
    \EndWhile
    \EndFunction
  \end{algorithmic}
\end{algorithm}

\section{Training Batch Distribution}
\label{sec:training_batch}

We use the same problem distribution for both training and evaluation since out-of-distribution generalization is not within the scope of this paper.
That is, when teaching 6-digit multiplication to the model, both training and test sets are all examples of 6-digit multiplication.
The problem distributions are elaborated in Appendix \ref{sec:prob_spec}.
Another important detail regarding the training of RoT is that each training example in a batch is a context, not a whole problem.
Since RoT generates multiple contexts per problem, often a large portion of contexts can be a duplicate (mostly the base cases). 
Therefore, to build a training batch for RoT, we first sample a top-level problem and find the set of unique RoT contexts from the problem.
Out of the unique contexts, we randomly sample one context as a training example.
We find this simple technique works well, and we do not need a more sophisticated method, such as the adaptive curriculum learning in \citet{Reed2016NeuralP}.

\section{Target Sequence}
\label{sec:target_seq}

\begin{algorithm}[H]
  \caption{Creating the target sequence}
  \label{alg:context_conversion}
  \begin{algorithmic}[1]
    \Require Context $X=[Q; Q^{\mathrm{sub},1}; A^{\mathrm{sub},1}; \allowbreak \ldots; Q^{\mathrm{sub},N}; A^{\mathrm{sub},N}; A]$
    \State $Y \gets \underbrace{\PAD ... \PAD}_{|Q|}$ \label{alg:ctx:q}
    \For{$n$ in $1...N$}
    \State $Y \gets [Y; Q^{\mathrm{sub},n}]$
    \State $Y \gets [Y; \THINK\underbrace{\PAD ... \PAD}_{|A^{\mathrm{sub},n}| - 1}]$ \label{alg:ctx:y-a-sub}
    \EndFor \label{alg:ctx:sub_end}
    \State $Y \gets [Y; A]$ \label{alg:ctx:a}
    \State \Return $Y$
  \end{algorithmic}
\end{algorithm}

\begin{figure*}[h]
  \centering
	\includegraphics[width=0.9\textwidth, trim={0 22.8cm 0 0}, clip]{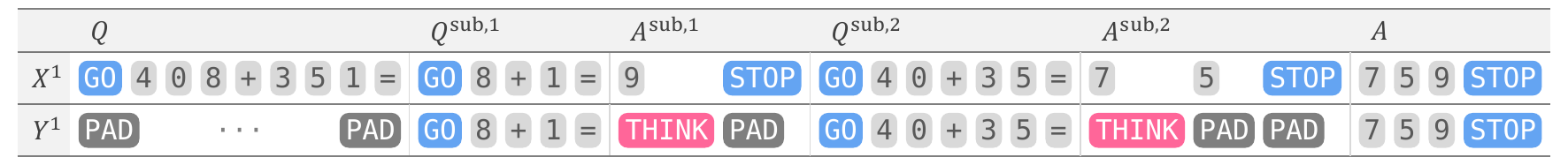}
	\caption{
		The target sequence $Y^1$ for $X^1$ in Figure \ref{fig:rot}.
	}
	\label{fig:target}
\end{figure*}

\section{A Step-by-Step Illustration of RoT Inference}
\label{sec:illustrated_rot}

In this section, we provide a step-by-step illustration of the example in Figure \ref{fig:rot}.
Here we assume an ideal model fully trained for RoT.

\newtcolorbox{stepbox}[2][]
{
  colframe = black!15!white,
  colback  = white,
  coltitle = black!90!white,  
  title    = {#2},
  #1
}

\begin{stepbox}{Step 1}
The context is initialized with the question $Q$.

\vspace{3mm}
\begin{tabular}{ll}
\toprule
& $Q$ \\
\midrule
$X^1$ & \cGO\tokenize{408+351=} \\
\bottomrule
\end{tabular}
\end{stepbox}

\begin{stepbox}{Step 2}
The model generates the first subquestion $8+1$.

\vspace{3mm}
\begin{tabular}{lll}
\toprule
& $Q$ & $Q^{\mathrm{sub},1}$ \\
\midrule
$X^1$ & \cGO\tokenize{408+351=} & \cGO\tokenize{8+1=} \\
\bottomrule
\end{tabular}
\end{stepbox}

\begin{stepbox}{Step 3}
Instead of immediately producing the answer, the model outputs the \cTHINK token.

\vspace{3mm}
\begin{tabular}{llll}
\toprule
& $Q$ & $Q^{\mathrm{sub},1}$ & $A^{\mathrm{sub},1}$ \\
\midrule
$X^1$ & \cGO\tokenize{408+351=} & \cGO\tokenize{8+1=} & \cTHINK \\
\bottomrule
\end{tabular}
\end{stepbox}

\begin{stepbox}{Step 4}
The \cTHINK token triggers the creation of a new context.
The new context is initialized with the subproblem starting from the last \cGO of $X^1$, i.e., $8+1$.

\vspace{3mm}
\begin{tabular}{llll}
\toprule
& $Q$ & $Q^{\mathrm{sub},1}$ & $A^{\mathrm{sub},1}$ \\
\midrule
$X^1$ & \cGO\tokenize{408+351=} & \cGO\tokenize{8+1=} & \cTHINK \\
\bottomrule
\end{tabular}

\vspace{5mm}

\begin{tabular}{ll}
\toprule
& $Q$ \\
\midrule
$X^2$ & \cGO\tokenize{8+1=} \\
\bottomrule
\end{tabular}
\end{stepbox}

\begin{stepbox}{Step 5}
Since the subproblem is a base case, the model outputs the answer $9$ immediately.

\vspace{3mm}
\begin{tabular}{llll}
\toprule
& $Q$ & $Q^{\mathrm{sub},1}$ & $A^{\mathrm{sub},1}$ \\
\midrule
$X^1$ & \cGO\tokenize{408+351=} & \cGO\tokenize{8+1=} & \cTHINK \\
\bottomrule
\end{tabular}

\vspace{5mm}

\begin{tabular}{lll}
\toprule
& $Q$ & $A$ \\
\midrule
$X^2$ & \cGO\tokenize{8+1=} & \tokenize{9}\cSTOP \\
\bottomrule
\end{tabular}
\end{stepbox}

\begin{stepbox}{Step 6}
The answer is returned and replaces the \cTHINK token.

\vspace{3mm}
\begin{tabular}{llll}
\toprule
& $Q$ & $Q^{\mathrm{sub},1}$ & $A^{\mathrm{sub},1}$ \\
\midrule
$X^1$ & \cGO\tokenize{408+351=} & \cGO\tokenize{8+1=} & \tokenize{9}\cSTOP \\
\bottomrule
\end{tabular}
\end{stepbox}

\begin{stepbox}{Step 7}
The model generates the next subproblem, which is to add the remaining digits.
Then, it produces \cTHINK to find its answer.

\vspace{3mm}
\begin{tabular}{llllll}
\toprule
& $Q$ & $Q^{\mathrm{sub},1}$ & $A^{\mathrm{sub},1}$ & $Q^{\mathrm{sub},2}$ & $A^{\mathrm{sub},2}$ \\
\midrule
$X^1$ & \cGO\tokenize{408+351=} & \cGO\tokenize{8+1=} & \tokenize{9}\cSTOP & \cGO\tokenize{40+35=} & \cTHINK  \\
\bottomrule
\end{tabular}
\end{stepbox}

\begin{stepbox}{Step 8}
The \cTHINK token creates a new context $X^3$ for solving $40+35$.

\vspace{3mm}
\begin{tabular}{llllll}
\toprule
& $Q$ & $Q^{\mathrm{sub},1}$ & $A^{\mathrm{sub},1}$ & $Q^{\mathrm{sub},2}$ & $A^{\mathrm{sub},2}$ \\
\midrule
$X^1$ & \cGO\tokenize{408+351=} & \cGO\tokenize{8+1=} & \tokenize{9}\cSTOP & \cGO\tokenize{40+35=} & \cTHINK  \\
\bottomrule
\end{tabular}
    
\vspace{5mm}

\begin{tabular}{ll}
\toprule
& $Q$ \\
\midrule
$X^3$ & \cGO\tokenize{40+35=} \\
\bottomrule
\end{tabular}
\end{stepbox}

\begin{stepbox}{Step 9}
Since $40+35$ is not a base case, the model recursively produces more subproblems.
In this case, the first subproblem is to add the last digits, i.e., 0 and 5.
Then it outputs the \cTHINK token to solve the subproblem.

\vspace{3mm}
\begin{tabular}{llllll}
\toprule
& $Q$ & $Q^{\mathrm{sub},1}$ & $A^{\mathrm{sub},1}$ & $Q^{\mathrm{sub},2}$ & $A^{\mathrm{sub},2}$ \\
\midrule
$X^1$ & \cGO\tokenize{408+351=} & \cGO\tokenize{8+1=} & \tokenize{9}\cSTOP & \cGO\tokenize{40+35=} & \cTHINK  \\
\bottomrule
\end{tabular}
    
\vspace{5mm}

\begin{tabular}{llll}
\toprule
& $Q$ & $Q^{\mathrm{sub},1}$ & $A^{\mathrm{sub},1}$ \\
\midrule
$X^3$ & \cGO\tokenize{40+35=} & \cGO\tokenize{0+5=} & \cTHINK \\
\bottomrule
\end{tabular}
\end{stepbox}

\begin{stepbox}{Step 10}
The new context $X^4$ is created to solve $0+5$.

\vspace{3mm}
\begin{tabular}{llllll}
\toprule
& $Q$ & $Q^{\mathrm{sub},1}$ & $A^{\mathrm{sub},1}$ & $Q^{\mathrm{sub},2}$ & $A^{\mathrm{sub},2}$ \\
\midrule
$X^1$ & \cGO\tokenize{408+351=} & \cGO\tokenize{8+1=} & \tokenize{9}\cSTOP & \cGO\tokenize{40+35=} & \cTHINK  \\
\bottomrule
\end{tabular}
    
\vspace{5mm}

\begin{tabular}{llll}
\toprule
& $Q$ & $Q^{\mathrm{sub},1}$ & $A^{\mathrm{sub},1}$ \\
\midrule
$X^3$ & \cGO\tokenize{40+35=} & \cGO\tokenize{0+5=} & \cTHINK \\
\bottomrule
\end{tabular}

\vspace{5mm}

\begin{tabular}{lll}
\toprule
& $Q$ & $A$ \\
\midrule
$X^4$ & \cGO\tokenize{0+5=} & \tokenize{5}\cSTOP \\
\bottomrule
\end{tabular}
\end{stepbox}

\begin{stepbox}{Step 11}
The answer is returned to $X^3$ and replaces the \cTHINK token.

\vspace{3mm}
\begin{tabular}{llllll}
\toprule
& $Q$ & $Q^{\mathrm{sub},1}$ & $A^{\mathrm{sub},1}$ & $Q^{\mathrm{sub},2}$ & $A^{\mathrm{sub},2}$ \\
\midrule
$X^1$ & \cGO\tokenize{408+351=} & \cGO\tokenize{8+1=} & \tokenize{9}\cSTOP & \cGO\tokenize{40+35=} & \cTHINK  \\
\bottomrule
\end{tabular}
    
\vspace{5mm}

\begin{tabular}{llll}
\toprule
& $Q$ & $Q^{\mathrm{sub},1}$ & $A^{\mathrm{sub},1}$ \\
\midrule
$X^3$ & \cGO\tokenize{40+35=} & \cGO\tokenize{0+5=} & \tokenize{5}\cSTOP \\
\bottomrule
\end{tabular}
\end{stepbox}

\begin{stepbox}{Step 12}
The model generates the next subproblem.

\vspace{3mm}
\begin{tabular}{llllll}
\toprule
& $Q$ & $Q^{\mathrm{sub},1}$ & $A^{\mathrm{sub},1}$ & $Q^{\mathrm{sub},2}$ & $A^{\mathrm{sub},2}$ \\
\midrule
$X^1$ & \cGO\tokenize{408+351=} & \cGO\tokenize{8+1=} & \tokenize{9}\cSTOP & \cGO\tokenize{40+35=} & \cTHINK  \\
\bottomrule
\end{tabular}
    
\vspace{5mm}

\begin{tabular}{llllll}
\toprule
& $Q$ & $Q^{\mathrm{sub},1}$ & $A^{\mathrm{sub},1}$ & $Q^{\mathrm{sub},2}$ & $A^{\mathrm{sub},2}$ \\
\midrule
$X^3$ & \cGO\tokenize{40+35=} & \cGO\tokenize{0+5=} & \tokenize{5}\cSTOP & \cGO\tokenize{4+3=} & \cTHINK \\
\bottomrule
\end{tabular}
\end{stepbox}

\begin{stepbox}{Step 13}
$X^5$ is created to solve the subproblem $4+3$.
Since this is a base case, the model produces the answer directly.

\vspace{3mm}
\begin{tabular}{llllll}
\toprule
& $Q$ & $Q^{\mathrm{sub},1}$ & $A^{\mathrm{sub},1}$ & $Q^{\mathrm{sub},2}$ & $A^{\mathrm{sub},2}$ \\
\midrule
$X^1$ & \cGO\tokenize{408+351=} & \cGO\tokenize{8+1=} & \tokenize{9}\cSTOP & \cGO\tokenize{40+35=} & \cTHINK  \\
\bottomrule
\end{tabular}
    
\vspace{5mm}

\begin{tabular}{llllll}
\toprule
& $Q$ & $Q^{\mathrm{sub},1}$ & $A^{\mathrm{sub},1}$ & $Q^{\mathrm{sub},2}$ & $A^{\mathrm{sub},2}$ \\
\midrule
$X^3$ & \cGO\tokenize{40+35=} & \cGO\tokenize{0+5=} & \tokenize{5}\cSTOP & \cGO\tokenize{4+3=} & \cTHINK \\
\bottomrule
\end{tabular}

\vspace{5mm}

\begin{tabular}{lll}
\toprule
& $Q$ & $A$ \\
\midrule
$X^5$ & \cGO\tokenize{4+3=} & \tokenize{7}\cSTOP \\
\bottomrule
\end{tabular}
\end{stepbox}

\begin{stepbox}{Step 14}
The answer from $X^5$ replaces the \cTHINK token in $X^3$.

\vspace{3mm}
\begin{tabular}{llllll}
\toprule
& $Q$ & $Q^{\mathrm{sub},1}$ & $A^{\mathrm{sub},1}$ & $Q^{\mathrm{sub},2}$ & $A^{\mathrm{sub},2}$ \\
\midrule
$X^1$ & \cGO\tokenize{408+351=} & \cGO\tokenize{8+1=} & \tokenize{9}\cSTOP & \cGO\tokenize{40+35=} & \cTHINK  \\
\bottomrule
\end{tabular}
    
\vspace{5mm}

\begin{tabular}{llllll}
\toprule
& $Q$ & $Q^{\mathrm{sub},1}$ & $A^{\mathrm{sub},1}$ & $Q^{\mathrm{sub},2}$ & $A^{\mathrm{sub},2}$ \\
\midrule
$X^3$ & \cGO\tokenize{40+35=} & \cGO\tokenize{0+5=} & \tokenize{5}\cSTOP & \cGO\tokenize{4+3=} & \tokenize{7}\cSTOP \\
\bottomrule
\end{tabular}
\end{stepbox}

\begin{stepbox}{Step 15}
Since all subproblems are solved in $X^3$, the answer $75$ is generated and returned to $X^1$.

\vspace{3mm}
\begin{tabular}{llllll}
\toprule
& $Q$ & $Q^{\mathrm{sub},1}$ & $A^{\mathrm{sub},1}$ & $Q^{\mathrm{sub},2}$ & $A^{\mathrm{sub},2}$ \\
\midrule
$X^1$ & \cGO\tokenize{408+351=} & \cGO\tokenize{8+1=} & \tokenize{9}\cSTOP & \cGO\tokenize{40+35=} & \cTHINK  \\
\bottomrule
\end{tabular}
    
\vspace{5mm}

\begin{tabular}{lllllll}
\toprule
& $Q$ & $Q^{\mathrm{sub},1}$ & $A^{\mathrm{sub},1}$ & $Q^{\mathrm{sub},2}$ & $A^{\mathrm{sub},2}$ & $A$ \\
\midrule
$X^3$ & \cGO\tokenize{40+35=} & \cGO\tokenize{0+5=} & \tokenize{5}\cSTOP & \cGO\tokenize{4+3=} & \tokenize{7}\cSTOP & \tokenize{75}\cSTOP \\
\bottomrule
\end{tabular}
\end{stepbox}

\begin{stepbox}{Step 16}
The answer of $X^3$ replaces the \cTHINK token in $X^1$.

\vspace{3mm}
\begin{tabular}{llllll}
\toprule
& $Q$ & $Q^{\mathrm{sub},1}$ & $A^{\mathrm{sub},1}$ & $Q^{\mathrm{sub},2}$ & $A^{\mathrm{sub},2}$ \\
\midrule
$X^1$ & \cGO\tokenize{408+351=} & \cGO\tokenize{8+1=} & \tokenize{9}\cSTOP & \cGO\tokenize{40+35=} & \tokenize{75}\cSTOP \\
\bottomrule
\end{tabular}
\end{stepbox}

\begin{stepbox}{Step 17}
Since the subproblems in $X^1$ are all solved, the model produces the final answer.

\vspace{3mm}
\begin{tabular}{llllll}
\toprule
& $Q$ & $Q^{\mathrm{sub},1}$ & $A^{\mathrm{sub},1}$ & $Q^{\mathrm{sub},2}$ & $A^{\mathrm{sub},2}$ \\
\midrule
$X^1$ & \cGO\tokenize{408+351=} & \cGO\tokenize{8+1=} & \tokenize{9}\cSTOP & \cGO\tokenize{40+35=} & \tokenize{75}\cSTOP \\
\bottomrule
\end{tabular}

\begin{flushright}
\begin{tabular}{l}
\toprule
$A$ \\
\midrule
\tokenize{759}\cSTOP \\
\bottomrule
\end{tabular}
\end{flushright}
\end{stepbox}

\section{Examples of CoT Training Data}
\label{sec:cot_example}

If we solve the example of 408+351 in Figure \ref{fig:rot} with RoT, the following five contexts are produced.
\begin{itemize}
	\item $X^1$: \tokenize{\GO408+351=\GO8+1=9\STOP\GO40+35=75\STOP759\STOP}
	\item $X^2$: \tokenize{\GO8+1=9\STOP}
	\item $X^3$: \tokenize{\GO40+35=\GO0+5=5\STOP\GO4+3=7\STOP75\STOP}
	\item $X^4$: \tokenize{\GO0+5=5\STOP}
	\item $X^5$: \tokenize{\GO4+3=7\STOP}
\end{itemize}

The CoT context of the same problem is:
\begin{itemize}
	\item $X^\mathrm{CoT}$: 
		\GO\tokenize{408+351=}
			\GO\tokenize{8+1=9}\STOP
			\GO\tokenize{40+35=}
				\GO\tokenize{0+5}\STOP
				\GO\tokenize{4+3}\STOP
			\tokenize{75}\STOP
		\tokenize{759}\STOP
\end{itemize}

In a slightly more complicated example of 34 $\times$ 5, the RoT contexts are as follows:
\begin{itemize}
\item $X^1$: \tokenize{\GO34*5=\GO4*5=20\STOP\GO3*5=15\STOP\TAIL150+20=\THINK}
\item $X^2$: \tokenize{\GO4*5=20\STOP}
\item $X^3$: \tokenize{\GO3*5=15\STOP}
\item $X^4$: \tokenize{\GO150+20=\GO0+0=0\STOP\GO15+2=17\STOP170\STOP}
\item $X^5$: \tokenize{\GO0+0=0\STOP}
\item $X^6$: \tokenize{\GO15+2=\GO5+2=7\STOP17\STOP}
\item $X^7$: \tokenize{\GO5+2=7\STOP}
\end{itemize}

The corresponding CoT context is:
\begin{itemize}
	\item $X^\mathrm{CoT}$: \tokenize{\GO34*5=\GO4*5=20\STOP\GO3*5=15\STOP\TAIL150+20=\GO0+0=0\STOP\GO15+2=\GO5+2=7\STOP17\STOP170\STOP}
\end{itemize}

Notice that the CoT context consists of all the corresponding RoT contexts as its subsequences.
The number of tokens to generate is identical to that of RoT if we do not count the \THINK tokens.
Even in these simple examples, however, the context size of CoT is far longer than that of RoT.
For much more complex problems, such as 8-digit multiplication or 0-1 Knapsack, the CoT context size can be orders of magnitude larger than RoT.
See Appendix \ref{sec:ctx_length} for more details on the distribution of context sizes.

\section{Problem Specifications}
\label{sec:prob_spec}

\subsection{The Arithmetic Problems}
\label{sec:prob_spec:arith}

For arithmetic tasks, we test addition, subtraction, multiplication, and division on non-negative integers.
For subtraction, we add a constraint that the first operand is not less than the second one, to enforce non-negative answers.
For division, we let the output include both a quotient and a remainder, separated by a special token \RMD, e.g., \GO\token{7}\DIV\tokenize{3=2}\RMD\token{1}\STOP.

As briefly mentioned in \S\ref{sec:probs}, naively sampling the operands from a uniform distribution makes the operands extremely biased towards large numbers.
For example, the probability of sampling a 2-digit number from the 6-digit space is less than 0.01\%.
Thus, we define a variation of the log-uniform distribution (often called the reciprocal distribution) to sample the operands.
As a result, we obtain roughly the same proportion of operands for each number of digits.

The probability density of a log-uniform distribution is proportional to the reciprocal of the value.
By definition, zero is not in the support of a log-uniform distribution, and samples are overly concentrated to the first few values in the sampling range.
Therefore, we slightly extend the log-uniform distribution by introducing an offset parameter $\delta$.
To sample an integer in range $[\alpha,\beta)$ with offset $\delta$, we first uniformly sample a real number $r$ in range $[\log (\alpha + \delta), \log (\beta + \delta)]$.
Then, $r$ is transformed to $\lfloor \exp(r) - \delta \rfloor$.
\newcommand{\logu}{U_\mathrm{log}}
We denote the extended log-uniform distribution $\logu(\alpha, \beta, \delta)$.
As $\delta$ gets larger, the samples are more dispersed to larger numbers.
In the experiments, we set $\delta = 3$.

Additionally, we introduce several other sampling details for division problems.
Assume that we independently sample two numbers $a$ and $b$ for the dividend and the divisor.
In about half of the cases, the dividend $a$ would be less than the divisor $b$, so the quotients will be zero for those cases.
To ensure a diverse range of quotients, we sample the divisor $b$ from $\logu(1, 10^N, 3)$, the quotient $c$ from $\logu(0, 10^N/b, 3)$, and the remainder $r$ from $\logu(0, b, 3)$.
The dividend is calculated from these values: $a=b\times c+r$.
This way, we can sample division problems with a diverse range of quotients and remainders.

Table \ref{tab:prob_samples} presents 40 problem samples for each 6-digit problem type.
Several properties of our sampling scheme can be observed in the table.
First, each number ranges over diverse numbers of digits.
Second, the division problems are mostly non-trivial, i.e., the quotients are not concentrated at zero.

\begin{table*}
	\centering
	\small
	\begin{tabular}{l|l|l|l}
		\toprule
		Addition & Subtraction & Multiplication & Division \\
		\midrule
		$1330+121163$ & $376776-35241$ & $9466\times176175$ & $620261\div155034$ \\
		$114780+4356$ & $10638-100$ & $179\times516$ & $111730\div1176$ \\
		$638+2$ & $109033-52649$ & $5509\times133$ & $28268\div1$ \\
		$35+77$ & $85137-3098$ & $6783\times2$ & $588137\div25571$ \\
		$114261+354$ & $22355-2824$ & $6\times80285$ & $180330\div739$ \\
		$3+13792$ & $7-1$ & $37275\times19258$ & $879975\div97772$ \\
		$10151+7$ & $652781-78853$ & $168484\times154$ & $111461\div905026$ \\
		$22+1399$ & $64914-3114$ & $3331\times40$ & $42338\div14003$ \\
		$363356+450475$ & $13041-1422$ & $349\times158$ & $108\div384103$ \\
		$73+11$ & $28293-4540$ & $17988\times262130$ & $60002\div7479$ \\
		$179895+4128$ & $11553-3576$ & $8140\times1670$ & $131467\div131290$ \\
		$3+10$ & $656291-2795$ & $51\times5$ & $890679\div62$ \\
		$1+141972$ & $93-42$ & $16497\times158$ & $228\div131108$ \\
		$57612+18403$ & $55972-1782$ & $74\times10$ & $892\div124$ \\
		$9+1621$ & $84587-51$ & $216\times13414$ & $15\div964156$ \\
		$3370+381$ & $273269-5867$ & $621\times2$ & $369044\div28364$ \\
		$678+8854$ & $274405-14$ & $2\times5951$ & $457\div46$ \\
		$422+10348$ & $51926-9$ & $189486\times13080$ & $14687\div730$ \\
		$118+582$ & $4272-229$ & $552792\times763$ & $200361\div1049$ \\
		$1343+408534$ & $223267-377$ & $77\times3$ & $19715\div965179$ \\
		$24+9251$ & $14857-1994$ & $179090\times469029$ & $98\div7$ \\
		$315+652424$ & $914771-836$ & $1037\times258$ & $406\div9$ \\
		$355+4434$ & $3035-2963$ & $8\times769974$ & $47345\div122$ \\
		$22+834928$ & $30-12$ & $47765\times7254$ & $391613\div1631$ \\
		$3028+357$ & $149-4$ & $5608\times18164$ & $892642\div3898$ \\
		$777+1355$ & $89057-6$ & $21437\times12$ & $241554\div1901$ \\
		$154874+81059$ & $296410-9$ & $15007\times15$ & $116475\div12908$ \\
		$64936+216852$ & $45-3$ & $539860\times427$ & $488317\div197443$ \\
		$3+340939$ & $78906-3$ & $3583\times9754$ & $7519\div325$ \\
		$3+984775$ & $56560-29960$ & $13\times66$ & $3560\div847611$ \\
		$50581+1183$ & $98-6$ & $266394\times185$ & $9711\div1385$ \\
		$415+943$ & $16551-920$ & $3988\times12$ & $44540\div103$ \\
		$110+49$ & $25606-194$ & $5514\times57$ & $19721\div58$ \\
		$15+17058$ & $45-37$ & $5\times1712$ & $59544\div24$ \\
		$36278+100$ & $129443-70196$ & $17\times430178$ & $333057\div333057$ \\
		$6+23516$ & $221-54$ & $227\times127$ & $25719\div5142$ \\
		$1462+848$ & $11010-818$ & $20888\times54$ & $7544\div46$ \\
		$1002+2773$ & $47759-67$ & $96\times232801$ & $45\div410$ \\
		$135+178346$ & $10-8$ & $175\times1050$ & $195659\div2047$ \\
		$22672+162038$ & $1439-153$ & $146\times166$ & $412572\div16$ \\
		\bottomrule
	\end{tabular}
	\caption{40 randomly selected samples of each type of 6-digit arithmetic problems.}
	\label{tab:prob_samples}
\end{table*}


\subsection{The Algorithmic Problems}
\label{sec:prob_spec:alg}

\subsubsection{Longest Common Subsequence (LCS)}
\label{sec:prob_spec:alg:lcs}

The question of an LCS problem is two number sequences joined by the \token{LCS} token, and the answer is the corresponding LCS and its length separated by \token{;}.
Here is an example of a length-4 LCS problem:
\begin{itemize}
	\item $Q$: \tokenize{\GO 1234}\token{LCS}\tokenize{2468=}
	\item $A$: \tokenize{24;2\STOP}
\end{itemize}
For a length-$N$ LCS problem, we sample two sequences of length $N$.
Each character of the sequences is randomly sampled from 0-9 with equal probability.

\subsubsection{Longest Palindromic Subsequence (LPS)}
\label{sec:prob_spec:alg:lps}

The question of a length-$N$ LPS problem starts with the \token{LPS}, followed by a sequence of length $N$.
Similar to LCS, the answer contains the corresponding LPS and its length separated by \token{;}.
The following is an example of a length-8 LPS problem:
\begin{itemize}
	\item $Q$: \GO\token{LPS}\tokenize{41253261=}
	\item $A$: \tokenize{12321;5\STOP}
\end{itemize}
The sequence of an LPS problem is sampled in the same way as done for the LCS problem.

\subsubsection{0-1 Knapsack}

Each item in a 0-1 Knapsack problem is represented by its value and weight.
For instance, \tokenize{12}\token{\&}\tokenize{34} represents an item with a value of 12 and a weight of 34.
The question part of a 0-1 Knapsack problem is a sequence consisting of the \token{KNAPSACK} token, a list of items separated by \token{,}, the token \token{@}, and the capacity of the knapsack.
The answer part starts with a list of items to include, then \token{\$}, and finally the total value.
The following is an example of a 3-item knapsack problem.
\begin{itemize}
	\item $Q$: \GO\token{KNAPSACK}\token{5}\token{\&}\tokenize{12,25}\token{\&}\tokenize{15,19}\token{\&}\tokenize{18@40=}
	\item $A$: \tokenize{25}\token{\&}\tokenize{15,19}\token{\&}\tokenize{18}\token{\$}\tokenize{44}\STOP
\end{itemize}
In this example, given a knapsack of capacity 40, the last two are selected with a total value of 44.

For a fixed number of items, we uniformly sample each item's value and weight from the integers of range [1, 99].

\subsubsection{Matrix Chain Multiplication (MCM)}

The cost of multiplying many matrices is very sensitive to the order of multiplication.
Matrix chain multiplication is the task of finding the best order with the minimum cost.
Here, the cost is defined to be the total number of element multiplications.
In the example of three matrices $A$, $B$, and $C$, whose shapes are $4\times2$, $2\times8$, and $8\times3$ respectively,
the cost of computing $(AB)C$ is $4\times2\times8 + 4\times8\times3=160$, while another order $A(BC)$ costs only $2\times8\times3 + 4\times2\times3=72$.
In the question of an MCM problem, the sizes of the matrices are enumerated, and the answer contains the order and the total cost separated by \token{;}.
The example above is represented as the following sequences.
\begin{itemize}
	\item $Q$: \GO\token{MCM}\tokenize{4}\TIMES\tokenize{2,2}\TIMES\tokenize{8,8}\TIMES\tokenize{3=}
	\item $A$: \tokenize{4}\TIMES\tokenize{2,(2}\TIMES\tokenize{8,8}\TIMES\tokenize{3);72}\STOP
\end{itemize}
Given a fixed number of matrices, we sample the sizes of matrices from the range [1, 99].


\subsubsection{Sorting}

Although not included in the main text, we test the problem of sorting multi-digit numbers.
The results are presented in Appendix \ref{sec:sorting}.
The problem difficulty is defined by the maximum number of terms.
For a sorting problem of at most $N$ terms, we first uniformly sample the number of terms from $[2, N]$.
Then we sample each term from $\logu (0, 1000, 5)$.
The following is an example of the sorting problem.
\begin{itemize}
\item $Q$: \GO\token{SORT}\tokenize{139,160,434,796,41=}
\item $A$: \tokenize{41,139,160,434,796}\STOP
\end{itemize}

\section{Details of the Recursive Reasoning Procedures}
\label{sec:data_gen}

\newcommand{\smtt}[1]{{\small\texttt{#1}}}

In this section, we elaborate on the procedures to recursively solve the arithmetic problems.
Specifically, we present the algorithms to produce the subproblems of a problem.
Therefore, for a set of randomly sampled questions, we can generate ground truth contexts using these algorithms.
For better understanding, we present the key parts of our Python code, the \smtt{thought} methods.
For each problem, we create a child class the \smtt{Problem} class and implement \smtt{thought} static method.
The method takes a set of arguments for a problem and returns the list of direct subproblems.
Each subproblem is represented by a problem class, problem arguments, and recursion type (whether it is a tail recursion or not).
We use named tuple \smtt{T} to group this information:
\begin{lstlisting}[language=Python]
from collections import namedtuple
T = namedtuple('Thought', ['prob_cls', 'args', 'type'], defaults=[''])
\end{lstlisting}
For instance, \smtt{T(Mul, (3, 4))} represents a regular subproblem of $3\times4$, and \smtt{T(Add, (12, 340), 'tail')} represents a subproblem of $12+340$ which should be performed as a tail recursion.
Once the \smtt{thought} method returns a list of \smtt{T}s, we can recursively find more subproblems for each subproblem.

\subsection{Addition}
The core idea of our recursive procedure for addition is to first add the last digits and then add the rest.
If the sum of the last digits is greater than or equal to 10, we insert another subproblem for adding the carry right after adding the last digits.

\begin{lstlisting}[language=Python]
class Add(Problem):
    @staticmethod
    def thought(args) -> list[T]:
        left, right = args

        # Base cases
        if left < 10 and right < 10:
            return []

        l_last, r_last = left % 10, right % 10
        thoughts = [T(Add, (l_last, r_last))]

        l_rest, r_rest = left // 10, right // 10
        if l_last + r_last >= 10:
            thoughts.append(T(Add, (l_rest, 1)))
            l_rest += 1

        if l_rest > 0 and r_rest > 0:
            thoughts.append(T(Add, (l_rest, r_rest)))

        return thoughts
\end{lstlisting}

Figure \ref{fig:rot} in the main draft is an example with no carry, and the following is another example of 27+65 with a carry.

\begin{itemize}
\item $X^{1}$: \cGO\token{3}\token{1}\token{7}\token{+}\token{6}\token{5}\token{=}\cGO\token{7}\token{+}\token{5}\token{=}\token{1}\token{2}\cSTOP\cGO\token{3}\token{1}\token{+}\token{1}\token{=}\token{3}\token{2}\cSTOP\cGO\token{3}\token{2}\token{+}\token{6}\token{=}\token{3}\token{8}\cSTOP\token{3}\token{8}\token{2}\cSTOP
\item $X^{2}$: \cGO\token{7}\token{+}\token{5}\token{=}\token{1}\token{2}\cSTOP
\item $X^{3}$: \cGO\token{3}\token{1}\token{+}\token{1}\token{=}\cGO\token{1}\token{+}\token{1}\token{=}\token{2}\cSTOP\token{3}\token{2}\cSTOP
\item $X^{4}$: \cGO\token{1}\token{+}\token{1}\token{=}\token{2}\cSTOP
\item $X^{5}$: \cGO\token{3}\token{2}\token{+}\token{6}\token{=}\cGO\token{2}\token{+}\token{6}\token{=}\token{8}\cSTOP\token{3}\token{8}\cSTOP
\item $X^{6}$: \cGO\token{2}\token{+}\token{6}\token{=}\token{8}\cSTOP
\end{itemize}

\subsection{Subtraction}

\newcommand{\tab}{\-\hspace{4.2mm}}
\newcommand{\context}[1]{\vspace{3mm}\noindent\hspace{4mm}\parbox{\textwidth}{#1}\vspace{2mm}}
Similar to addition, we first subtract the last digits and solve the rest recursively.
When subtracting the last digits $x$ and $y$, we always borrow 10 for $x$ to prevent a negative result.
The borrowing of 10 is easy for a sequence model: just put \token{1} before $x$.
Therefore, the base cases of subtraction are when $a \leq 19$ and $b \leq 9$.
If the subtraction result of the last digits is smaller than 10, i.e., the borrow is actually needed, we subtract 1 from the rest of the first operand $m$.

\begin{lstlisting}[language=Python]
class Sub(Problem):
    @staticmethod
    def thought(args) -> list[T]:
        left, right = args

        # Base cases
        if left <= 19 and right <= 9:
            return []

        l_last = left % 10 + 10
        r_last = right % 10
        thoughts = [T(Sub, (l_last, r_last))]
        l_rest, r_rest = left // 10, right // 10
        if l_last - r_last < 10:
            thoughts.append(T(Sub, (l_rest, 1)))
            l_rest -= 1
        if r_rest > 0:
            thoughts.append(T(Sub, (l_rest, r_rest)))

        return thoughts
\end{lstlisting}

Here is an example of 432-216:

\begin{itemize}
\item $X^{1}$: \cGO\token{4}\token{3}\token{2}\token{-}\token{2}\token{1}\token{6}\token{=}\cGO\token{1}\token{2}\token{-}\token{6}\token{=}\token{6}\cSTOP\cGO\token{4}\token{3}\token{-}\token{1}\token{=}\token{4}\token{2}\cSTOP\cGO\token{4}\token{2}\token{-}\token{2}\token{1}\token{=}\token{2}\token{1}\cSTOP\token{2}\token{1}\token{6}\cSTOP
\item $X^{2}$: \cGO\token{1}\token{2}\token{-}\token{6}\token{=}\token{6}\cSTOP
\item $X^{3}$: \cGO\token{4}\token{3}\token{-}\token{1}\token{=}\cGO\token{1}\token{3}\token{-}\token{1}\token{=}\token{1}\token{2}\cSTOP\token{4}\token{2}\cSTOP
\item $X^{4}$: \cGO\token{1}\token{3}\token{-}\token{1}\token{=}\token{1}\token{2}\cSTOP
\item $X^{5}$: \cGO\token{4}\token{2}\token{-}\token{2}\token{1}\token{=}\cGO\token{1}\token{2}\token{-}\token{1}\token{=}\token{1}\token{1}\cSTOP\cGO\token{4}\token{-}\token{2}\token{=}\token{2}\cSTOP\token{2}\token{1}\cSTOP
\item $X^{6}$: \cGO\token{1}\token{2}\token{-}\token{1}\token{=}\token{1}\token{1}\cSTOP
\item $X^{7}$: \cGO\token{4}\token{-}\token{2}\token{=}\token{2}\cSTOP
\end{itemize}
Notice that the final answer and the questions of each subproblem can be easily constructed from the previous sequence.

\subsection{Multiplication}

The base cases of multiplication are (i) when either operand is 0 or 1, or (ii) when both operands are less than 10.
If one of the operands is 0, then the answer is zero; when one of them is 1, then the answer is just a copy of the other operand.
For the cases where both operands are less than 10, we just let the model memorize them, which is similar to an elementary school math curriculum.

There are two types of non-base cases.
For the simpler case, where the second operand is less than 10, we first split the first operand into the last digit and the rest.
We then multiply each of them with the second operand and combine the results.
Otherwise, we split the second operand into the last digit and the rest.
The first operand is multiplied to each of them, and the results are summed.

\begin{lstlisting}[language=Python]
class Mul(Problem):
    @staticmethod
    def thought(args) -> list[T]:
        left, right = args

        # Base cases
        if left <= 1 or right <= 1:
            return []
        if left <= 9 and right <= 9:
            return []

        thoughts = []
        if right < 10:
            thoughts.append(T(Mul, (left % 10, right)))
            thoughts.append(T(Mul, (left // 10, right)))

            a1 = (left % 10) * right
            a2 = (left // 10) * right
            thoughts.append(T(Add, (a2 * 10, a1), 'tail'))
        else:
            a1 = left * (right % 10)
            thoughts.append(T(Mul, (left, right % 10)))

            a2 = left * (right // 10)
            thoughts.append(T(Mul, (left, right // 10)))

            thoughts.append(T(Add, (a2 * 10, a1), 'tail'))
        return thoughts
\end{lstlisting}

Here are some example contexts of multiplication:

\begin{itemize}
\item $X^{1}$: \cGO\token{4}\token{3}\token{*}\token{2}\token{1}\token{=}\cGO\token{4}\token{3}\token{*}\token{1}\token{=}\token{4}\token{3}\cSTOP\cGO\token{4}\token{3}\token{*}\token{2}\token{=}\token{8}\token{6}\cSTOP\cTAIL\token{8}\token{6}\token{0}\token{+}\token{4}\token{3}\token{=}\cTHINK
\item $X^{2}$: \cGO\token{4}\token{3}\token{*}\token{1}\token{=}\token{4}\token{3}\cSTOP
\item $X^{3}$: \cGO\token{4}\token{3}\token{*}\token{2}\token{=}\cGO\token{3}\token{*}\token{2}\token{=}\token{6}\cSTOP\cGO\token{4}\token{*}\token{2}\token{=}\token{8}\cSTOP\cTAIL\token{8}\token{0}\token{+}\token{6}\token{=}\cTHINK
\item $X^{4}$: \cGO\token{3}\token{*}\token{2}\token{=}\token{6}\cSTOP
\item $X^{5}$: \cGO\token{4}\token{*}\token{2}\token{=}\token{8}\cSTOP
\item $X^{6}$: \cGO\token{8}\token{0}\token{+}\token{6}\token{=}\cGO\token{0}\token{+}\token{6}\token{=}\token{6}\cSTOP\token{8}\token{6}\cSTOP
\item $X^{7}$: \cGO\token{0}\token{+}\token{6}\token{=}\token{6}\cSTOP
\item $X^{8}$: \cGO\token{8}\token{6}\token{0}\token{+}\token{4}\token{3}\token{=}\cGO\token{0}\token{+}\token{3}\token{=}\token{3}\cSTOP\cGO\token{8}\token{6}\token{+}\token{4}\token{=}\token{9}\token{0}\cSTOP\token{9}\token{0}\token{3}\cSTOP
\item $X^{9}$: \cGO\token{0}\token{+}\token{3}\token{=}\token{3}\cSTOP
\item $X^{10}$: \cGO\token{8}\token{6}\token{+}\token{4}\token{=}\cGO\token{6}\token{+}\token{4}\token{=}\token{1}\token{0}\cSTOP\cGO\token{8}\token{+}\token{1}\token{=}\token{9}\cSTOP\token{9}\token{0}\cSTOP
\item $X^{11}$: \cGO\token{6}\token{+}\token{4}\token{=}\token{1}\token{0}\cSTOP
\item $X^{12}$: \cGO\token{8}\token{+}\token{1}\token{=}\token{9}\cSTOP
\end{itemize}
Notice that we use tail recursion in $X^1$ and $X^3$.

\subsection{Comparison}

Comparison is used as a subroutine during division.
The procedure for comparison consists of three steps:
\begin{enumerate}
	\item Compare the numbers of digits.
	\item If the numbers of digits are the same, compare the most significant digits.
	\item If the most significant digits are identical, compare the remaining digits recursively.
\end{enumerate}
We find that the sequence models can perform the first step without an explicit subproblem.
Therefore, we only add intermediate steps for the second and third steps.

\begin{lstlisting}[language=Python]
class Compare(Problem):
    @staticmethod
    def thought(args) -> list[T]:
        left, right = args

        # Base cases
        if left < 10 and right < 10:
            return []

        thoughts = []
        digit_l, digit_r = len(str(left)), len(str(right))
        if digit_l == digit_r:
            # Compare the first digits
            l_first, r_first = int(str(left)[0]), int(str(right)[0])
            thoughts.append(T(Compare, (l_first, r_first)))
            if l_first == r_first:
                # Compare the rest
                l_rest = int(str(left)[1:])
                r_rest = int(str(right)[1:])
                thoughts.append(T(Compare, (l_rest, r_rest)))

        return thoughts
\end{lstlisting}

The following is an example of comparing 153 and 159.
\begin{itemize}
\item $X^{1}$: \cGO\token{1}\token{5}\token{3}\token{VS}\token{1}\token{5}\token{9}\token{=}\cGO\token{1}\token{VS}\token{1}\token{=}\token{EQ}\cSTOP\cGO\token{5}\token{3}\token{VS}\token{5}\token{9}\token{=}\token{LT}\cSTOP\token{LT}\cSTOP
\item $X^{2}$: \cGO\token{1}\token{VS}\token{1}\token{=}\token{EQ}\cSTOP
\item $X^{3}$: \cGO\token{5}\token{3}\token{VS}\token{5}\token{9}\token{=}\cGO\token{5}\token{VS}\token{5}\token{=}\token{EQ}\cSTOP\cGO\token{3}\token{VS}\token{9}\token{=}\token{LT}\cSTOP\token{LT}\cSTOP
\item $X^{4}$: \cGO\token{5}\token{VS}\token{5}\token{=}\token{EQ}\cSTOP
\item $X^{5}$: \cGO\token{3}\token{VS}\token{9}\token{=}\token{LT}\cSTOP
\end{itemize}

\subsection{Division}

Solving division is the most challenging among the four basic arithmetic operations
since the procedure is basically trial and error, searching for the correct quotient.
Nonetheless, the following process is a recursive version of the elementary school division.

The base case is when the dividend is less than or equal to the divisor.
If the dividend is smaller than the divisor, the quotient is 0, and the remainder is the dividend.
If the dividend is equal to the divisor, then the quotient is 1, and the remainder is 0.
Both cases can be handled relatively easily by neural sequence models.
To determine whether it is one of these cases, we always perform the comparison as the first subproblem.

If it is not a base case, we check whether the dividend is smaller than 10 times the divisor.
If the dividend is smaller, we subtract the divisor from the dividend and recursively divide the result with the divisor.
The final answer is attained by simply adding 1 to the quotient of the smaller division.

To explain the other case, where the dividend is greater than 10 times the divisor, let us call the dividend $a$ and the divisor $b$.
First, we split the $a$ into the last digit $x$ and the remaining digits $m$.
Then, we divide $m$ with the divisor $b$, i.e., we are solving the one-digit-smaller subproblem first.
Since we define the division operation to return both a quotient and a remainder, the quotient $q_1=m/b$ and the remainder $r_1=m \mod b$ from the subproblem are added to the context.
Next, we \textit{concatenate} the remainder and $x$, which is numerically computing $r \times 10 + x$, and divide it again with $b$.
Let the quotient and the remainder of this operation $q_2$ and $r_2$.
Then, the quotient of the final answer is $q_1 \times 10 + q_2$, while the remainder is simply $r_2$.

\begin{lstlisting}[language=Python]
class Div(Problem):
    @staticmethod
    def thought(args) -> list[T]:
        left, right = args
        thoughts = [T(Compare, (left, right))]

        # Base cases
        if left <= right:
            return thoughts

        thoughts.append(T(Compare, (left, right * 10)))
        if left <= right * 10:
            diff = left - right
            thoughts.append(T(Sub, (left, right)))
            thoughts.append(T(Div, (diff, right)))
        else:
            thoughts.append(T(Div, (left // 10, right)))
            left_remainder = (left // 10) % right * 10 + left % 10
            thoughts.append(T(Div, (left_remainder, right)))
        return thoughts
\end{lstlisting}

The following is an example of $76\div29$.
\begin{itemize}
\item $X^{1}$: \cGO\token{7}\token{6}\token{÷}\token{2}\token{9}\token{=}\cGO\token{7}\token{6}\token{VS}\token{2}\token{9}\token{=}\token{GT}\cSTOP\cGO\token{7}\token{6}\token{VS}\token{2}\token{9}\token{0}\token{=}\token{LT}\cSTOP\cGO\token{7}\token{6}\token{-}\token{2}\token{9}\token{=}\token{4}\token{7}\cSTOP\cGO\token{4}\token{7}\token{÷}\token{2}\token{9}\token{=}\token{1}\token{R}\token{1}\token{8}\cSTOP\token{2}\token{R}\token{1}\token{8}\cSTOP
\item $X^{2}$: \cGO\token{7}\token{6}\token{VS}\token{2}\token{9}\token{=}\cGO\token{7}\token{VS}\token{2}\token{=}\token{GT}\cSTOP\token{GT}\cSTOP
\item $X^{3}$: \cGO\token{7}\token{VS}\token{2}\token{=}\token{GT}\cSTOP
\item $X^{4}$: \cGO\token{7}\token{6}\token{VS}\token{2}\token{9}\token{0}\token{=}\token{LT}\cSTOP
\item $X^{5}$: \cGO\token{7}\token{6}\token{-}\token{2}\token{9}\token{=}\cGO\token{1}\token{6}\token{-}\token{9}\token{=}\token{7}\cSTOP\cGO\token{7}\token{-}\token{1}\token{=}\token{6}\cSTOP\cGO\token{6}\token{-}\token{2}\token{=}\token{4}\cSTOP\token{4}\token{7}\cSTOP
\item ...
\item $X^{9}$: \cGO\token{4}\token{7}\token{÷}\token{2}\token{9}\token{=}\cGO\token{4}\token{7}\token{VS}\token{2}\token{9}\token{=}\token{GT}\cSTOP\cGO\token{4}\token{7}\token{VS}\token{2}\token{9}\token{0}\token{=}\token{LT}\cSTOP\cGO\token{4}\token{7}\token{-}\token{2}\token{9}\token{=}\token{1}\token{8}\cSTOP\cGO\token{1}\token{8}\token{÷}\token{2}\token{9}\token{=}\token{0}\token{R}\token{1}\token{8}\cSTOP\token{1}\token{R}\token{1}\token{8}\cSTOP
\item $X^{10}$: \cGO\token{4}\token{7}\token{VS}\token{2}\token{9}\token{=}\cGO\token{4}\token{VS}\token{2}\token{=}\token{GT}\cSTOP\token{GT}\cSTOP
\item $X^{11}$: \cGO\token{4}\token{VS}\token{2}\token{=}\token{GT}\cSTOP
\item $X^{12}$: \cGO\token{4}\token{7}\token{VS}\token{2}\token{9}\token{0}\token{=}\token{LT}\cSTOP
\item $X^{13}$: \cGO\token{4}\token{7}\token{-}\token{2}\token{9}\token{=}\cGO\token{1}\token{7}\token{-}\token{9}\token{=}\token{8}\cSTOP\cGO\token{4}\token{-}\token{1}\token{=}\token{3}\cSTOP\cGO\token{3}\token{-}\token{2}\token{=}\token{1}\cSTOP\token{1}\token{8}\cSTOP
\item ...
\item $X^{17}$: \cGO\token{1}\token{8}\token{÷}\token{2}\token{9}\token{=}\cGO\token{1}\token{8}\token{VS}\token{2}\token{9}\token{=}\token{LT}\cSTOP\token{0}\token{R}\token{1}\token{8}\cSTOP
\item $X^{18}$: \cGO\token{1}\token{8}\token{VS}\token{2}\token{9}\token{=}\cGO\token{1}\token{VS}\token{2}\token{=}\token{LT}\cSTOP\token{LT}\cSTOP
\item ...
\end{itemize}

\subsection{Longest Common Subsequence (LCS)}

Given sequences $A$ and $B$, the algorithm starts by comparing the last characters of the two sequences.
If the last two characters are the same, we find LCS of the subsequences without the last characters, i.e., LCS of $A_{:-1}$ and $B_{:-1}$.
Otherwise, we compute the LCSs of the cases where the last character of either side is removed and return the better one.
In the following code, \smtt{LCS.\_answer} is the subroutine that finds the LCS of two sequences.
\smtt{Equal} returns \token{TRUE} if the two arguments are the same, or \token{FALSE} otherwise.

\begin{lstlisting}[language=Python]
class LCS(Problem):
    @staticmethod
    def thought(args) -> list[T]:
        l, r = args
        if len(l) == 0 or len(r) == 0:
            return []

        thoughts = [T(Equal, (l[-1], r[-1]))]
        if l[-1] == r[-1]:
            thoughts.append(T(LCS, (l[:-1], r[:-1])))
            return thoughts

        lcs1_args = (l[:-1], r)
        lcs2_args = (l, r[:-1])
        lcs1 = LCS._answer(lcs1_args)
        lcs2 = LCS._answer(lcs2_args)
        thoughts.extend([
            T(LCS, lcs1_args),
            T(LCS, lcs2_args),
            T(Compare, (len(lcs1), len(lcs2)))
        ])
        return thoughts
\end{lstlisting}

The following is an example of finding the LCS of \smtt{123} and \smtt{234}.
\begin{itemize}
\item $X^{1}$: \cGO\token{1}\token{2}\token{3}\token{LCS}\token{2}\token{3}\token{4}\token{=}\cGO\token{EQUAL}\token{3}\token{,}\token{4}\token{=}\token{FALSE}\cSTOP\cGO\token{1}\token{2}\token{LCS}\token{2}\token{3}\token{4}\token{=}\token{2}\token{;}\token{1}\cSTOP\cGO\token{1}\token{2}\token{3}\token{LCS}\token{2}\token{3}\token{=}\token{2}\token{3}\token{;}\token{2}\cSTOP\cGO\token{1}\token{VS}\token{2}\token{=}\token{LT}\cSTOP\token{2}\token{3}\token{;}\token{2}\cSTOP
\item $X^{2}$: \cGO\token{EQUAL}\token{3}\token{,}\token{4}\token{=}\token{FALSE}\cSTOP
\item $X^{3}$: \cGO\token{1}\token{2}\token{LCS}\token{2}\token{3}\token{4}\token{=}\cGO\token{EQUAL}\token{2}\token{,}\token{4}\token{=}\token{FALSE}\cSTOP\cGO\token{1}\token{LCS}\token{2}\token{3}\token{4}\token{=}\token{;}\token{0}\cSTOP\cGO\token{1}\token{2}\token{LCS}\token{2}\token{3}\token{=}\token{2}\token{;}\token{1}\cSTOP\cGO\token{0}\token{VS}\token{1}\token{=}\token{LT}\cSTOP\token{2}\token{;}\token{1}\cSTOP
\item ...
\item $X^{21}$: \cGO\token{1}\token{2}\token{3}\token{LCS}\token{2}\token{3}\token{=}\cGO\token{EQUAL}\token{3}\token{,}\token{3}\token{=}\token{TRUE}\cSTOP\cGO\token{1}\token{2}\token{LCS}\token{2}\token{=}\token{2}\token{;}\token{1}\cSTOP\token{2}\token{3}\token{;}\token{2}\cSTOP
\item ...
\item $X^{23}$: \cGO\token{1}\token{VS}\token{2}\token{=}\token{LT}\cSTOP
\end{itemize}

\subsection{Longest Palindromic Subsequence (LPS)}

The overall algorithm for LPS is similar to LCS.
The base cases are when the sequence length is less than 3.
If it is not a base case, we first check if the characters at both ends of the sequence are the same.
If they are the same, we find the LPS of the subsequence excluding them.
Otherwise, we compare the cases where one of the end characters is excluded.

\begin{lstlisting}[language=Python]
class LPS(Problem):
    @staticmethod
    def thought(args) -> list[T]:
        # Base cases
        if len(args) == 1:
            return []
        elif len(args) == 2:
            return [T(Equal, args)]

        thoughts = [T(Equal, (args[0], args[1]))]
        if args[0] == args[-1]:
            sub_lps = LPS._answer(args[1:-1])
            thoughts.extend([
                T(LPS, args[1:-1]),
                T(Add, (len(sub_lps), 2))
            ])
        else:
            lps1_args = args[:-1]
            lps2_args = args[1:]
            lps1 = LPS._answer(lps1_args)
            lps2 = LPS._answer(lps2_args)
            thoughts.extend([
                T(LPS, lps1_args),
                T(LPS, lps2_args),
                T(Compare, (len(lps1), len(lps2)))
            ])
        return thoughts
\end{lstlisting}

The following is an example of LPS.

\begin{itemize}
\item $X^{1}$: \cGO\token{LPS}\token{1}\token{2}\token{3}\token{2}\token{=}\cGO\token{EQUAL}\token{1}\token{,}\token{2}\token{=}\token{FALSE}\cSTOP\cGO\token{LPS}\token{1}\token{2}\token{3}\token{=}\token{1}\token{;}\token{1}\cSTOP\cGO\token{LPS}\token{2}\token{3}\token{2}\token{=}\token{2}\token{3}\token{2}\token{;}\token{3}\cSTOP\cGO\token{1}\token{VS}\token{3}\token{=}\token{LT}\cSTOP\token{2}\token{3}\token{2}\token{;}\token{3}\cSTOP
\item $X^{2}$: \cGO\token{EQUAL}\token{1}\token{,}\token{2}\token{=}\token{FALSE}\cSTOP
\item $X^{3}$: \cGO\token{LPS}\token{1}\token{2}\token{3}\token{=}\cGO\token{EQUAL}\token{1}\token{,}\token{3}\token{=}\token{FALSE}\cSTOP\cGO\token{LPS}\token{1}\token{2}\token{=}\token{1}\token{;}\token{1}\cSTOP\cGO\token{LPS}\token{2}\token{3}\token{=}\token{2}\token{;}\token{1}\cSTOP\cGO\token{1}\token{VS}\token{1}\token{=}\token{EQ}\cSTOP\token{1}\token{;}\token{1}\cSTOP
\item ...
\item $X^{10}$: \cGO\token{LPS}\token{2}\token{3}\token{2}\token{=}\cGO\token{EQUAL}\token{2}\token{,}\token{2}\token{=}\token{TRUE}\cSTOP\cGO\token{LPS}\token{3}\token{=}\token{3}\token{;}\token{1}\cSTOP\cGO\token{1}\token{+}\token{2}\token{=}\token{3}\cSTOP\token{2}\token{3}\token{2}\token{;}\token{3}\cSTOP
\item ...
\item $X^{14}$: \cGO\token{1}\token{VS}\token{3}\token{=}\token{LT}\cSTOP
\end{itemize}

\subsection{0-1 Knapsack}

The base cases are when there is only one item.
In this case, we simply compare the item's weight and the knapsack's capacity, to determine whether the item should be included.
If it is a non-base case, we compare two possibilities: (i) include the first item, or (ii) exclude the first item.
We recursively compute the subproblems and find the case with the best value.

\begin{lstlisting}[language=Python]
class LPS(Problem):
    @staticmethod
    def thought(args) -> list[T]:
        items, capacity = args
        value, weight = items[0]

        # Base case
        if len(items) == 1:
            return [T(Compare, (weight, capacity))]

        # When excluding the current item
        items_max, value_max = Knapsack._answer((items[1:], capacity))
        thoughts = [
            T(Knapsack, (items[1:], capacity)),
            T(Compare, (weight, capacity)),
        ]

        # When including the current item
        if weight <= capacity:
            items_sub, value_sub = Knapsack._answer(
                (items[1:], capacity - weight))
            value_incl = value_sub + value
            thoughts.extend([
                T(Sub, (capacity, weight)),
                T(Knapsack, (items[1:], capacity - weight)),
                T(Add, (value_sub, value)),
                T(Compare, (value_incl, value_max)),
            ])

        return thoughts
\end{lstlisting}

The following is an example of a 0-1 knapsack problem with three items and a knapsack capacity of 10.
\begin{itemize}
\item $X^{1}$: \cGO\token{KNAPSACK}\token{3}\token{\&}\token{9}\token{,}\token{4}\token{\&}\token{2}\token{,}\token{9}\token{\&}\token{5}\token{@}\token{1}\token{0}\token{=}\cGO\token{KNAPSACK}\token{4}\token{\&}\token{2}\token{,}\token{9}\token{\&}\token{5}\token{@}\token{1}\token{0}\token{=}\token{4}\token{\&}\token{2}\token{,}\token{9}\token{\&}\token{5}\token{\$}\token{1}\token{3}\cSTOP\cGO\token{9}\token{VS}\token{1}\token{0}\token{=}\token{LT}\cSTOP\cGO\token{1}\token{0}\token{-}\token{9}\token{=}\token{1}\cSTOP\cGO\token{KNAPSACK}\token{4}\token{\&}\token{2}\token{,}\token{9}\token{\&}\token{5}\token{@}\token{1}\token{=}\token{\$}\token{0}\cSTOP\cGO\token{0}\token{+}\token{3}\token{=}\token{3}\cSTOP\cGO\token{3}\token{VS}\token{1}\token{3}\token{=}\token{LT}\cSTOP\token{4}\token{\&}\token{2}\token{,}\token{9}\token{\&}\token{5}\token{\$}\token{1}\token{3}\cSTOP
\item $X^{2}$: \cGO\token{KNAPSACK}\token{4}\token{\&}\token{2}\token{,}\token{9}\token{\&}\token{5}\token{@}\token{1}\token{0}\token{=}\cGO\token{KNAPSACK}\token{9}\token{\&}\token{5}\token{@}\token{1}\token{0}\token{=}\token{9}\token{\&}\token{5}\token{\$}\token{9}\cSTOP\cGO\token{2}\token{VS}\token{1}\token{0}\token{=}\token{LT}\cSTOP\cGO\token{1}\token{0}\token{-}\token{2}\token{=}\token{8}\cSTOP\cGO\token{KNAPSACK}\token{9}\token{\&}\token{5}\token{@}\token{8}\token{=}\token{9}\token{\&}\token{5}\token{\$}\token{9}\cSTOP\cGO\token{9}\token{+}\token{4}\token{=}\token{1}\token{3}\cSTOP\cGO\token{1}\token{3}\token{VS}\token{9}\token{=}\token{GT}\cSTOP\token{4}\token{\&}\token{2}\token{,}\token{9}\token{\&}\token{5}\token{\$}\token{1}\token{3}\cSTOP
\item ...
\item $X^{11}$: \cGO\token{9}\token{VS}\token{1}\token{0}\token{=}\token{LT}\cSTOP
\item $X^{12}$: \cGO\token{1}\token{0}\token{-}\token{9}\token{=}\token{1}\cSTOP
\item $X^{13}$: \cGO\token{KNAPSACK}\token{4}\token{\&}\token{2}\token{,}\token{9}\token{\&}\token{5}\token{@}\token{1}\token{=}\cGO\token{KNAPSACK}\token{9}\token{\&}\token{5}\token{@}\token{1}\token{=}\token{\$}\token{0}\cSTOP\cGO\token{2}\token{VS}\token{1}\token{=}\token{GT}\cSTOP\token{\$}\token{0}\cSTOP
\item ...
\item $X^{17}$: \cGO\token{0}\token{+}\token{3}\token{=}\token{3}\cSTOP
\item $X^{18}$: \cGO\token{3}\token{VS}\token{1}\token{3}\token{=}\token{LT}\cSTOP
\end{itemize}

\subsection{Ternary Addition and Multiplication}

Ternary addition and multiplication arise as subproblems while solving MCM, which will be explained in the next section.
They are simple extensions of addition and multiplication to three integers.
\begin{lstlisting}[language=Python]
class TernaryAdd(Problem):
    @staticmethod
    def thought(args) -> list[T]:
        a1, a2, a3 = args
        return [
            T(Add, (a1, a2)),
            T(Add, (a1 + a2, a3), 'tail')
        ]


class TernaryMul(Problem):
    @staticmethod
    def thought(args) -> list[T]:
        a1, a2, a3 = args
        return [
            T(Mul, (a1, a2)),
            T(Mul, (a1 * a2, a3), 'tail')
        ]
\end{lstlisting}

\subsection{Matrix Chain Multiplication (MCM)}

Given $N$ matrices, the $N-1$ subproblems are defined for each possible binary split.
For the multiplication of four matrices $ABCD$, there are three possible binary splits: $A(BCD)$, $(AB)(CD)$, and $(ABC)D$.
For each binary split, the total cost is the sum of (i) the minimum cost of computing the first group, (ii) the minimum cost of computing the second group, and (iii) the cost of multiplying the two matrices resulting from each group.
Once we get the total costs of each binary split, we return the best split with the minimum cost.
The following code implements this procedure.

\begin{lstlisting}[language=Python]
class MCM(Problem):
    @staticmethod
    def thought(args) -> list[T]:
        mats, min_order, min_cost = args

        # Base cases
        if len(mats) == 1:
            return []

        if min_order is None:
            # Top-level problem
            l_mats, r_mats = mats[:1], mats[1:]
        else:
            # Middle of recursion
            l_mats, r_mats = mats

        l_args = (l_mats, None, None)
        r_args = (r_mats, None, None)
        l_order, l_cost = MCM._answer(l_args)
        r_order, r_cost = MCM._answer(r_args)
        agg_cost = l_mats[0][0] * r_mats[0][0] * r_mats[-1][1]
        thoughts = [
            T(MCM, l_args),
            T(MCM, r_args),
            T(TernaryMul, (l_mats[0][0], r_mats[0][0], r_mats[-1][1])),
            T(TernaryAdd, (l_cost, r_cost, agg_cost)),
        ]

        cost = l_cost + r_cost + agg_cost
        if min_cost is not None:
            thoughts.append(T(Compare, (cost, min_cost)))
        if min_cost is None or cost < min_cost:
            min_cost = cost
            min_order = l_order, r_order

        if len(r_mats) > 1:
            new_l_mats = l_mats + (r_mats[0],)
            new_r_mats = r_mats[1:]
            thoughts.append(
                T(MCM, ((new_l_mats, new_r_mats), min_order, min_cost), 'tail'))

        return thoughts
\end{lstlisting}

The following is an example of a three-matrix MCM.
\begin{itemize}
\item $X^{1}$: \cGO\token{MCM}\token{3}\token{×}\token{9}\token{,}\token{9}\token{×}\token{4}\token{,}\token{4}\token{×}\token{5}\token{=}\cGO\token{MCM}\token{3}\token{×}\token{9}\token{=}\token{3}\token{×}\token{9}\token{;}\token{0}\cSTOP\cGO\token{MCM}\token{9}\token{×}\token{4}\token{,}\token{4}\token{×}\token{5}\token{=}\token{9}\token{×}\token{4}\token{,}\token{4}\token{×}\token{5}\token{;}\token{1}\token{8}\token{0}\cSTOP\cGO\token{3}\token{*}\token{9}\token{*}\token{5}\token{=}\token{1}\token{3}\token{5}\cSTOP\cGO\token{0}\token{+}\token{1}\token{8}\token{0}\token{+}\token{1}\token{3}\token{5}\token{=}\token{3}\token{1}\token{5}\cSTOP\cTAIL\token{MCM}\token{3}\token{×}\token{9}\token{,}\token{9}\token{×}\token{4}\token{|}\token{4}\token{×}\token{5}\token{ACC}\token{3}\token{×}\token{9}\token{,}\token{(}\token{9}\token{×}\token{4}\token{,}\token{4}\token{×}\token{5}\token{)}\token{;}\token{3}\token{1}\token{5}\token{=}\cTHINK
\item ...
\item $X^{32}$: \cGO\token{MCM}\token{3}\token{×}\token{9}\token{,}\token{9}\token{×}\token{4}\token{|}\token{4}\token{×}\token{5}\token{ACC}\token{3}\token{×}\token{9}\token{,}\token{(}\token{9}\token{×}\token{4}\token{,}\token{4}\token{×}\token{5}\token{)}\token{;}\token{3}\token{1}\token{5}\token{=}\cGO\token{MCM}\token{3}\token{×}\token{9}\token{,}\token{9}\token{×}\token{4}\token{=}\token{3}\token{×}\token{9}\token{,}\token{9}\token{×}\token{4}\token{;}\token{1}\token{0}\token{8}\cSTOP\cGO\token{MCM}\token{4}\token{×}\token{5}\token{=}\token{4}\token{×}\token{5}\token{;}\token{0}\cSTOP\cGO\token{3}\token{*}\token{4}\token{*}\token{5}\token{=}\token{6}\token{0}\cSTOP\cGO\token{1}\token{0}\token{8}\token{+}\token{0}\token{+}\token{6}\token{0}\token{=}\token{1}\token{6}\token{8}\cSTOP\cGO\token{1}\token{6}\token{8}\token{VS}\token{3}\token{1}\token{5}\token{=}\token{LT}\cSTOP\token{(}\token{3}\token{×}\token{9}\token{,}\token{9}\token{×}\token{4}\token{)}\token{,}\token{4}\token{×}\token{5}\token{;}\token{1}\token{6}\token{8}\cSTOP
\item ...
\end{itemize}

\subsection{Sorting}

Among several sorting algorithms, we choose merge sort for our experiments with CoT and RoT.
Note that WT is not relevant to the sorting algorithm since it produces the answer directly.
The merge sort algorithm is simple: (i) split the given sequence into two equally sized subsequences, (ii) sort each subsequence, and (iii) merge the two sorted sequences.
Since the final merge operation is quite complicated, we define the merge as a problem type.

\begin{lstlisting}[language=Python]
class Merge(Problem):
    @staticmethod
    def thought(args) -> list[T]:
        l, r = args
        if len(l) == 0 or len(r) == 0:
            return []

        thoughts = [T(Compare, (l[0], r[0]))]
        if l[0] < r[0] and len(l) > 1:
            thoughts.append(T(Merge, (l[1:], r)))
        elif l[0] >= r[0] and len(r) > 1:
            thoughts.append(T(Merge, (l, r[1:])))
        return thoughts


class MergeSort(Problem):
    @staticmethod
    def thought(args) -> list[T]:
        if len(args) < 2:
            return []

        l_len = (len(args) + 1) // 2
        l = args[:l_len]
        r = args[l_len:]
        return [
            T(MergeSort, l),
            T(MergeSort, r),
            T(Merge, (tuple(sorted(l)), tuple(sorted(r))), 'tail')
        ]
\end{lstlisting}

\section{Fine-Tuning GPT-3 for Recursion of Thought}
\label{sec:gpt-3}

Using the OpenAI API, we fine-tune GPT-3 for Recursion of Thought.
The goal is to learn 16-digit addition, 16-digit subtraction, 8-digit multiplication, and 8-digit division simultaneously.
GPT-3's fine-tuning API takes a dataset where each example is a prompt-completion pair in plain text.
It is converted to tokens by a special tokenizer for GPT, which we cannot control.
This API is not directly compatible with RoT due to several reasons.
\begin{itemize}
	\item There is no special tokens such as \GO, \THINK, and \STOP.
	\item The input and target sequences have to be the same.
	However, they are different in RoT due to the \THINK token.
	Once \THINK is produced, the RoT framework triggers the recursion process to find the subproblem's answer and replace the \THINK token with it.
	Therefore, the \THINK token appears in the target sequences, but never in the input sequences.
\end{itemize}

Moreover, the way that GPT-3 tokenizes numbers hinders the learning of arithmetic reasoning rules.
GPT-3 tokenizes a multi-digit number into a set of two-digit or three-digit numbers.
For example, the text \texttt{1234567} is converted to the sequence of tokens \token{123}\token{45}\token{67}.
Under this tokenization scheme, the relationship between the numbers becomes obscured.
As an example, the tokens \token{7}, \token{17}, \token{27}, ..., \token{997} all have 7 as their last digit.
Since there is no direct way for a model to know that they share the same digit,
it is crucial to use each digit as a token.
We believe that OpenAI needs to correct this tokenization of GPT-3 for numbers. 

Luckily, we can mimic the RoT procedures with the API by using several tricks.
First, we replace the special tokens with plain lower-case words, e.g., \GO $\to$ \texttt{go} and \STOP $\to$ \texttt{stop}, which are included in the vocabulary of GPT-3.
Second, we add a space before each token to make sure that the GPT tokenizer separates each token.
We also add space before each digit to prevent the tokenizer from grouping a number into 2-to-3-digit tokens.
Finally, to simulate the behavior of the \THINK and \STOP tokens, we derive multiple examples from each context, one for each \THINK or \STOP output.

As an example, context $X^3$ in Figure \ref{fig:rot} is converted to the following JSON lines for GPT-3 as follows:
{\setlength{\tabcolsep}{1mm}
	\begin{center}
		\begin{tabular}{r|llll}
			\toprule
			$X^3$ & \GO\tokenize{40+35=} & \GO\tokenize{0+5=5}\STOP     & \GO\tokenize{4+3=7}\STOP     & \tokenize{75}\STOP \\
			\midrule
			$Y^3$ & \PAD $\times 7$ & \GO\tokenize{0+5=}\THINK\PAD & \GO\tokenize{4+3=}\THINK\PAD & \tokenize{75}\STOP \\
			\bottomrule
		\end{tabular}
	\end{center}
}
$$
\Downarrow
$$
\lstdefinestyle{gpt-3}{
	basicstyle=\ttfamily\scriptsize,
	breakatwhitespace=false,         
	breaklines=true,                 
	captionpos=b,                    
	keepspaces=true,                 
	numbers=left,                    
	numbersep=5pt,                  
	showspaces=false,                
	showstringspaces=true,
	showtabs=false,                  
	tabsize=2
}
\lstset{showstringspaces=true}
\begin{lstlisting}[language=Python]
{"prompt": " go 4 0 + 3 5 =", "completion": " go 0 + 5 = think"}
{"prompt": " go 4 0 + 3 5 = go 0 + 5 = 5 stop", "completion": " go 4 + 3 = think"}
{"prompt": " go 4 0 + 3 5 = go 0 + 5 = 5 stop go 4 + 3 = 7 stop", "completion": " 7 5 stop"}
\end{lstlisting}

In the case of Without Thought (WT), each problem is simply converted into a single example:
{\setlength{\tabcolsep}{1mm}
	\begin{center}
		\begin{tabular}{r|ll}
			\toprule
			$X$ & \GO\tokenize{40+35=} & \tokenize{75}\STOP \\
			\midrule
			$Y$ & \PAD $\times 7$ & \tokenize{75}\STOP \\
			\bottomrule
		\end{tabular}
	\end{center}
}
$$
\Downarrow
$$
\begin{lstlisting}[language=Python]
{"prompt": " go 4 0 + 3 5 =", "completion": " 7 5 stop"}
\end{lstlisting}

In both cases of RoT and WT, we fine-tune GPT-3 for 10K steps with a batch size of 256.
Among the several variants of GPT-3, we use Ada which is offered at the lowest cost.
Note that RoT produces multiple contexts for each problem, and each RoT context is converted to multiple training examples.
For this reason, the GPT-3 fine-tuned for RoT encounters much fewer problems during training, although the numbers of training steps are the same.

\section{Training Details of the Tiny Models}
\label{sec:tiny_models}

In all experiments, we use a batch size of 256 and Adam optimizer \cite{Kingma2015AdamAM} with a learning rate of 0.001, i.e., the default learning rate in PyTorch.
We train the Transformers for 500K steps and decay the learning rate by half every 50K steps.
Since the LSTMs converge slower than the Transformers, we train them for 800K steps and decay the learning rate by half every 100K steps.
At every 20K steps, we evaluate the model on a test set of 30K problems, and if a model reaches a perfect accuracy of 1.0, we do not train the model further.
The models can be trained on a single GPU with 12GB memory.

\section{Efficient Evaluation of RoT}
\label{sec:eval}

At the problem scales that RoT is tested, solving a single problem can require hundreds of thousands of tokens.
Therefore, we have to develop an efficient evaluation algorithm.
Since the models are deterministic, we do not need to evaluate the same context multiple times.
Therefore, given a test set, we first find all unique contexts that should appear while solving the test set.
Note that there would be a lot of duplicate contexts, especially for the base cases.
These unique contexts can be evaluated in parallel.
We evaluate each context in a similar manner to the training.
For each context, we construct the target sequence, as we described in \S\ref{sec:training}, and check if the model outputs the target sequence correctly, given the GT context.
In the case of GPT-3, we convert the contexts and target sequences as described in Appendix \ref{sec:gpt-3} to feed into the API.
Once every context is individually evaluated, the next step is to aggregate the results.
In this aggregation stage, a (sub)problem is marked correct only if its subproblems are all marked correct.
This way, we efficiently evaluate RoT with minimal cost.

\section{Failure Case Analysis of GPT-3 with RoT}
\label{sec:failure}

\newcommand{\smtt}[1]{{\small\texttt{#1}}}
\newcommand{\wrong}[1]{{\color{codeMagenta}#1}}
In this section, we provide some failure cases of RoT in the GPT-3 experiments for further insights.
Although the model achieves an accuracy of over 99\% in almost every case, there are several problems where it makes mistakes.
Nonetheless, only a few characters are wrong in each context.

The most common type of error is repeating the same character a wrong number of times.
The following is the examples of this error type.

\begin{center}
\begin{tabular}{l|p{11cm}}
\toprule
Prompt & \smtt{ go 1 6 4 3 4 vs 1 0 1 1 6 = go 1 vs 1 = eq stop} \\
\midrule
Completion   & \smtt{ go 6 4 3 4 vs \wrong{1 1} 6 = think} \\
\midrule
Model Output & \smtt{ go 6 4 3 4 vs \wrong{1 1 1} 6 = think} \\
\bottomrule
\end{tabular}

\begin{tabular}{l|p{11cm}}
\toprule
Prompt & \smtt{ go 4 8 0 9 4 4 9 5 1 8 2 5 5 9 1 * 4 4 3 0 2 5 3 6 0 1 7 = go 4 8 0 9 4 4 9 5 1 8 2 5 5 9 1 * 7 = 3 3 6 6 6 1 4 6 6 2 7 7 9 1 3 7 stop go 4 8 0 9 4 4 9 5 1 8 2 5 5 9 1 * 4 4 3 0 2 5 3 6 0 1 = 2 1 3 0 7 0 8 1 0 4 7 0 8 0 9 6 0 5 1 7 0 3 1 9 1 stop} \\
\midrule
Completion   & \smtt{ tail 2 1 3 0 7 0 8 1 0 4 7 0 8 0 9 6 0 5 1 7 0 3 1 9 1 0 + 3 3 \wrong{6 6 6} 1 4 6 6 2 7 7 9 1 3 7 = think} \\
\midrule
Model Output & \smtt{ tail 2 1 3 0 7 0 8 1 0 4 7 0 8 0 9 6 0 5 1 7 0 3 1 9 1 0 + 3 3 \wrong{6 6 6 6} 1 4 6 6 2 7 7 9 1 3 7 = think} \\
\bottomrule
\end{tabular}

\begin{tabular}{l|p{11cm}}
\toprule
Prompt & \smtt{ go 1 0 5 4 6 2 0 * 1 5 = go 1 0 5 4 6 2 0 * 5 = 5 2 7 3 1 0 0 stop go 1 0 5 4 6 2 0 * 1 = 1 0 5 4 6 2 0 stop} \\
\midrule
Completion   & \smtt{ tail 1 0 5 4 6 2 \wrong{0 0} + 5 2 7 3 1 0 0 = think} \\
\midrule
Model Output & \smtt{ tail 1 0 5 4 6 2 \wrong{0 0 0} + 5 2 7 3 1 0 0 = think} \\
\bottomrule
\end{tabular}
\end{center}

Another common case is incorrectly counting the number of digits while comparing two numbers.
When comparing numbers following our recursive procedure, the model has to immediately return the answer if the numbers of digits are different.
In the following example, the first number has seven digits, and the second one has six digits.
\begin{center}
\begin{tabular}{l|p{11cm}}
\toprule
Prompt & \smtt{ go 1 0 6 9 0 3 4 vs 8 9 1 5 0 6 =} \\
\midrule
Completion   & \smtt{ \wrong{gt stop}} \\
\midrule
Model Output & \smtt{ \wrong{go 1 vs 8}} \\
\bottomrule
\end{tabular}
\end{center}
Although the model should directly output the answer, i.e., \smtt{ gt stop}, it decides that the numbers of digits are the same and moves on to the next step, which is to compare the most significant digits.
This error is caused by the specific recursive procedure of our choice.
Since we assume that the model would easily figure out the difference in the number of digits, we did not add explicit steps to count the digits.
Although the model detects the difference in most cases, it turns out to be a relatively challenging operation.
A possible solution is to teach the model to explicitly count the number of digits and compare them.

\section{Context Length Distribution}
\label{sec:ctx_length}

\begin{figure}
	\includegraphics[width=\textwidth]{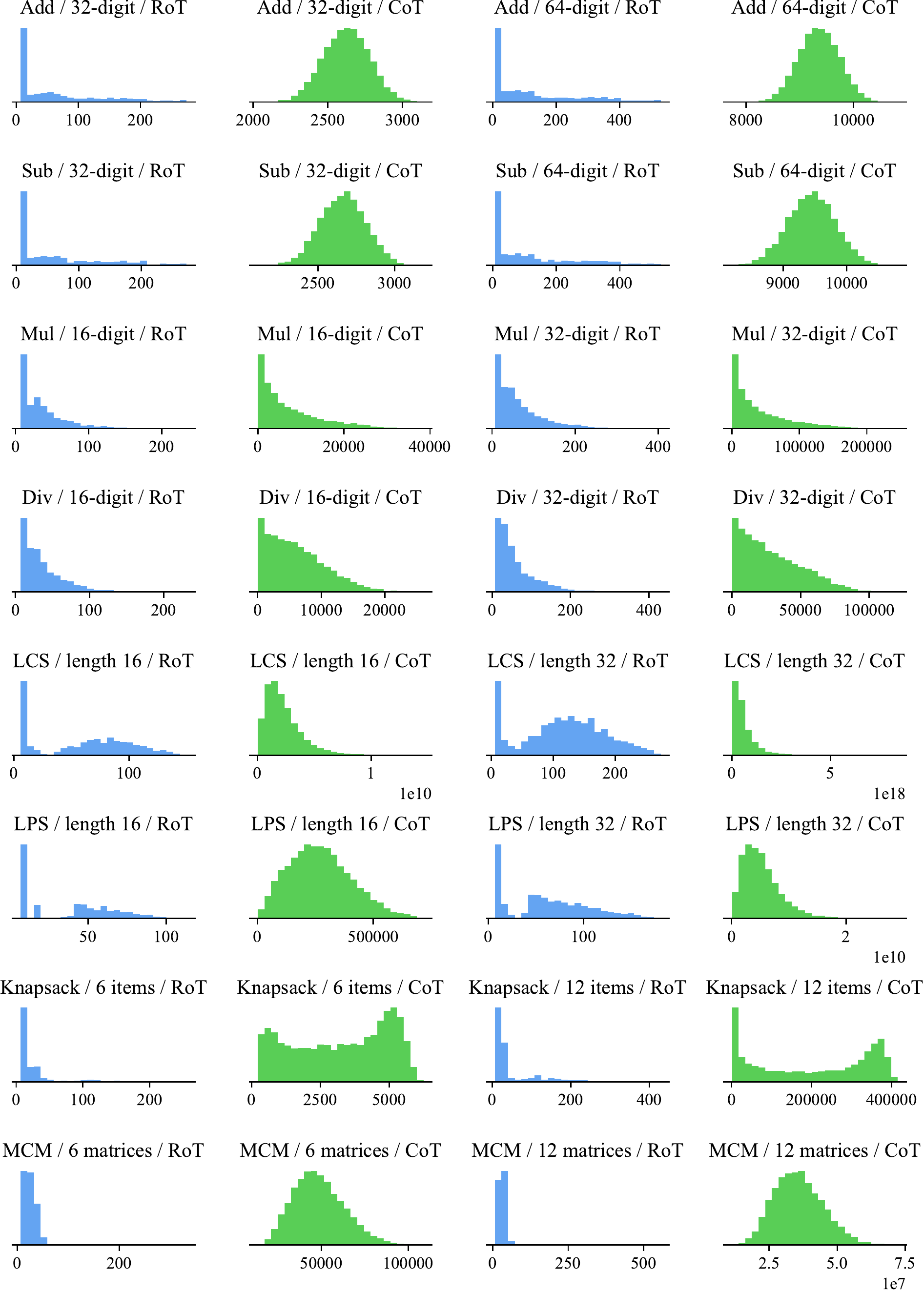}
	\caption{The distributions of context lengths.}
	\label{fig:ctx_length}
\end{figure}

In Figure \ref{fig:ctx_length}, we present the distributions of context lengths for each problem type.
We compare the context lengths of RoT and CoT.
For each configuration, we randomly sample 10K contexts from the training distribution and plot the histogram of their lengths.
The graphs show that the context sizes of CoT are many orders of magnitude larger than RoT.
In theory, the total number of tokens to generate for each problem is identical in both RoT and CoT (if we do not count the \THINK tokens).
However, RoT's context sizes are much smaller since it utilizes multiple contexts.

\begin{figure}
	\includegraphics[width=\textwidth]{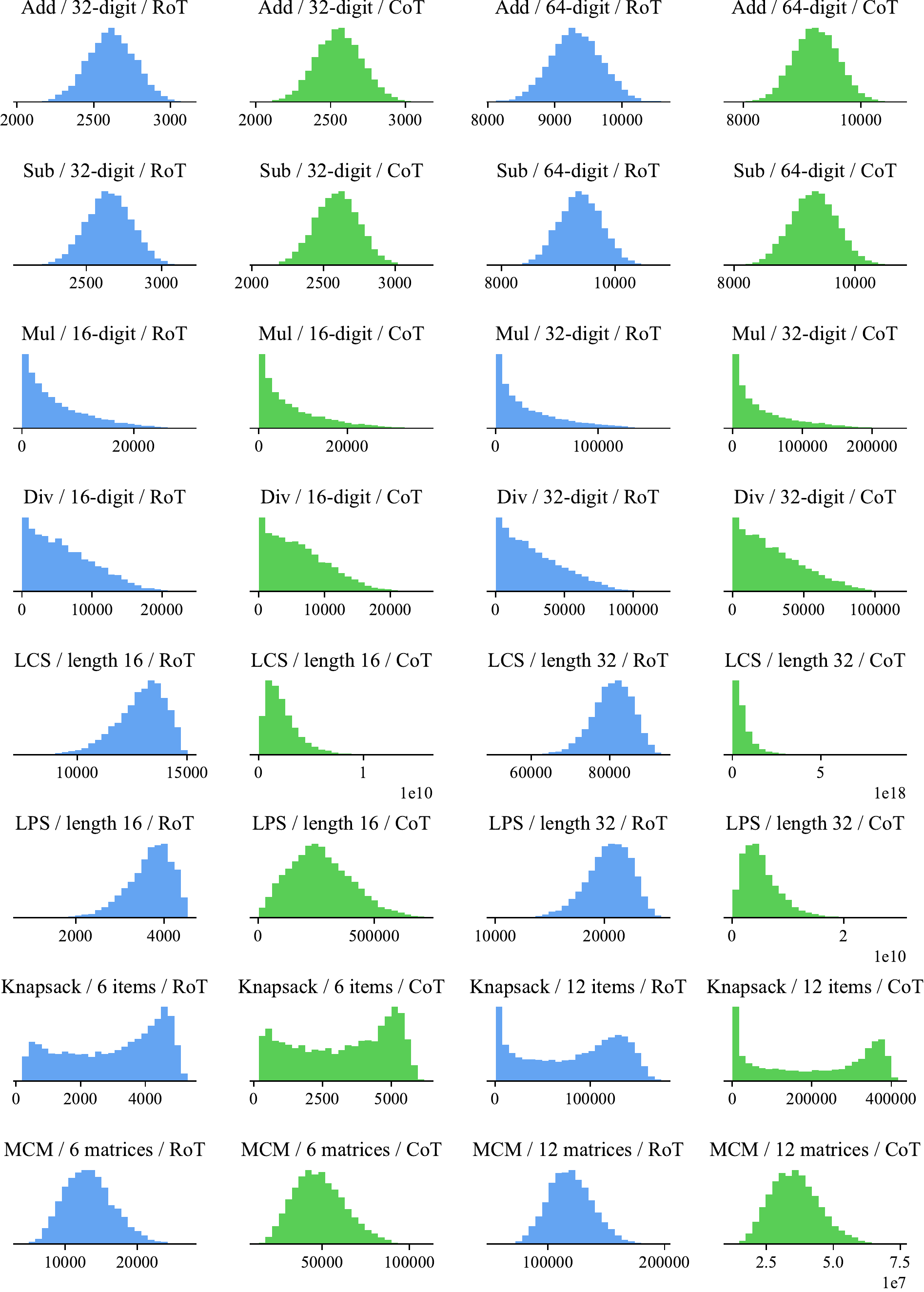}
	\caption{
	The distribution of the total number of tokens to produce in order to solve each problem.
	RoT can utilize dynamic programming to reduce redundant computations.
	}
	\label{fig:prob_tokens}
\end{figure}

Another advantage of RoT is the utilization of dynamic programming.
Since we can easily cache the duplicate computations of RoT as explained in Appendix \ref{sec:eval}, we can drastically reduce the amount of token generation if there is a redundant structure in the problem.
The amount of tokens to generate for each problem is plotted in Figure \ref{fig:prob_tokens}.
The benefit is especially prominent in algorithmic problems.
For example, finding the LCS of two 32-digit sequences results in more than $10^{18}$ tokens if we naively use CoT or RoT.
If we use dynamic programming with RoT, we can efficiently solve the same problem with much less cost.

\section{Transformers Are Powerful Sorting Machines}
\label{sec:sorting}

\begin{figure}
\centering
\includegraphics[width=0.6\textwidth]{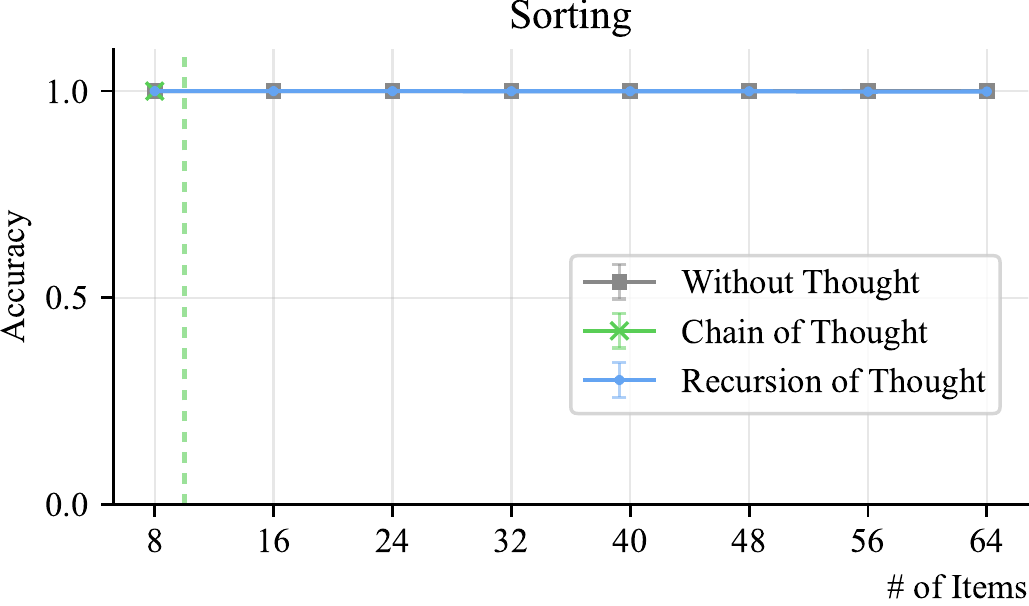}
\caption{Sorting experiment with the tiny Transformer.}
\label{fig:exp_sorting}
\end{figure}

In fact, the first algorithmic task that we tested is sorting since it has been widely used as a benchmark for algorithmic reasoning \citep{Reed2016NeuralP,Cai2017MakingNP,Pierrot2019LearningCN}.
However, we find that Transformers are incredibly good at sorting, even in the WT setting.
Figure \ref{fig:exp_sorting} shows the sorting experiment.
For CoT and RoT, we train the merge sort algorithm.
Interestingly, WT easily achieves a perfect score in sorting 64 three-digit numbers.
Also, the training converges much faster than RoT.
The Transformer architecture, more specifically the attention mechanism, seems to be perfectly suited for the sorting operation.

\section{The Exact Values of Figure \ref{fig:experiments}}
\label{sec:values}

Table \ref{tab:val:gpt-3}-\ref{tab:val:lstm} show the exact values of the graphs in Figure \ref{fig:experiments}.
Except for the GPT-3 experiments in Table \ref{tab:val:gpt-3}, we report the average and the standard deviation of eight runs.
Each GPT-3 experiment is done only once.

\begin{table}
\centering
\small
\begin{tabular}{l|l|rrr}
\toprule
Problem & Difficulty & WT & CoT & RoT \\
\midrule
\multirow{2}{*}{Addition}
& 32-digit & $0.991$ & $-$ & $0.998$ \\
& 48-digit & $0.853$ & $-$ & $0.995$ \\
\midrule
\multirow{2}{*}{Subtraction}
& 32-digit & $0.991$ & $-$ & $0.998$ \\
& 48-digit & $0.886$ & $-$ & $0.998$ \\
\midrule
\multirow{2}{*}{Multiplication}
& 8-digit & $0.337$ & $-$ & $0.999$ \\
& 16-digit & $0.098$ & $-$ & $0.994$ \\
\midrule
\multirow{2}{*}{Division}
& 8-digit & $0.363$ & $-$ & $1.000$ \\
& 16-digit & $0.123$ & $-$ & $0.989$ \\
\midrule
\multirow{2}{*}{LCS}
& length 16 & $0.980$ & $-$ & $0.995$ \\
& length 24 & $0.832$ & $-$ & $0.998$ \\
\midrule
\multirow{2}{*}{LPS}
& length 24 & $0.995$ & $-$ & $1.000$ \\
& length 40 & $0.800$ & $-$ & $0.974$ \\
\midrule
\multirow{2}{*}{0-1 Knapsack}
& 4 items & $0.945$ & $-$ & $0.999$ \\
& 6 items & $0.634$ & $-$ & $1.000$ \\
\midrule
\multirow{2}{*}{MCM}
& 3 matrices & $0.481$ & $-$ & $0.997$ \\
& 4 matrices & $0.110$ & $-$ & $0.992$ \\
\bottomrule
\end{tabular}
\caption{The exact values of the GPT-3 experiments in Figure \ref{fig:exp:gpt3}.}
\label{tab:val:gpt-3}
\end{table}

\begin{table}[h]
\centering
\small
\begin{tabular}{l|l|rrr}
\toprule
Problem & Difficulty & WT & CoT & RoT \\
\midrule
\multirow{8}{*}{Addition}
& 8-digit & $0.863 \pm 0.265$ & $1.000 \pm 0.000$ & $1.000 \pm 0.000$ \\
& 16-digit & $0.370 \pm 0.475$ & $1.000 \pm 0.000$ & $1.000 \pm 0.000$ \\
& 24-digit & $0.336 \pm 0.430$ & $1.000 \pm 0.000$ & $1.000 \pm 0.000$ \\
& 32-digit & $0.455 \pm 0.458$ & $-$ & $1.000 \pm 0.000$ \\
& 40-digit & $0.119 \pm 0.316$ & $-$ & $1.000 \pm 0.000$ \\
& 48-digit & $0.082 \pm 0.216$ & $-$ & $1.000 \pm 0.000$ \\
& 56-digit & $0.105 \pm 0.277$ & $-$ & $1.000 \pm 0.000$ \\
& 64-digit & $0.000 \pm 0.000$ & $-$ & $1.000 \pm 0.001$ \\
\midrule
\multirow{8}{*}{Subtraction}
& 8-digit & $0.982 \pm 0.006$ & $1.000 \pm 0.000$ & $1.000 \pm 0.000$ \\
& 16-digit & $0.705 \pm 0.411$ & $1.000 \pm 0.000$ & $1.000 \pm 0.000$ \\
& 24-digit & $0.238 \pm 0.412$ & $1.000 \pm 0.000$ & $1.000 \pm 0.000$ \\
& 32-digit & $0.221 \pm 0.385$ & $-$ & $1.000 \pm 0.000$ \\
& 40-digit & $0.426 \pm 0.433$ & $-$ & $1.000 \pm 0.000$ \\
& 48-digit & $0.114 \pm 0.303$ & $-$ & $1.000 \pm 0.000$ \\
& 56-digit & $0.116 \pm 0.307$ & $-$ & $1.000 \pm 0.000$ \\
& 64-digit & $0.161 \pm 0.282$ & $-$ & $1.000 \pm 0.000$ \\
\midrule
\multirow{9}{*}{Multiplication}
& 2-digit & $1.000 \pm 0.000$ & $1.000 \pm 0.000$ & $1.000 \pm 0.000$ \\
& 4-digit & $0.817 \pm 0.023$ & $1.000 \pm 0.000$ & $1.000 \pm 0.000$ \\
& 8-digit & $0.340 \pm 0.032$ & $-$ & $1.000 \pm 0.000$ \\
& 12-digit & $0.169 \pm 0.015$ & $-$ & $1.000 \pm 0.000$ \\
& 16-digit & $0.104 \pm 0.016$ & $-$ & $1.000 \pm 0.000$ \\
& 20-digit & $0.048 \pm 0.020$ & $-$ & $1.000 \pm 0.000$ \\
& 24-digit & $0.033 \pm 0.017$ & $-$ & $0.999 \pm 0.001$ \\
& 28-digit & $0.014 \pm 0.006$ & $-$ & $0.999 \pm 0.001$ \\
& 32-digit & $0.012 \pm 0.001$ & $-$ & $0.999 \pm 0.000$ \\
\midrule
\multirow{9}{*}{Division}
& 2-digit & $1.000 \pm 0.000$ & $1.000 \pm 0.000$ & $1.000 \pm 0.000$ \\
& 4-digit & $0.978 \pm 0.008$ & $1.000 \pm 0.000$ & $1.000 \pm 0.000$ \\
& 8-digit & $0.354 \pm 0.029$ & $-$ & $1.000 \pm 0.000$ \\
& 12-digit & $0.186 \pm 0.009$ & $-$ & $1.000 \pm 0.000$ \\
& 16-digit & $0.128 \pm 0.011$ & $-$ & $1.000 \pm 0.000$ \\
& 20-digit & $0.087 \pm 0.012$ & $-$ & $1.000 \pm 0.000$ \\
& 24-digit & $0.075 \pm 0.005$ & $-$ & $1.000 \pm 0.000$ \\
& 28-digit & $0.059 \pm 0.007$ & $-$ & $0.999 \pm 0.000$ \\
& 32-digit & $0.048 \pm 0.008$ & $-$ & $0.999 \pm 0.000$ \\
\bottomrule
\end{tabular}
\caption{The exact values of the Transformer experiments in Figure \ref{fig:exp:transformer} (arithmetic problems).}
\label{tab:val:transformer-arith}
\end{table}

\begin{table}[h]
\centering
\small
\begin{tabular}{l|l|rrr}
\toprule
Problem & Difficulty & WT & CoT & RoT \\
\midrule
\multirow{9}{*}{LCS}
& length 3 & $1.000 \pm 0.000$ & $1.000 \pm 0.000$ & $-$ \\
& length 4 & $0.997 \pm 0.008$ & $-$ & $1.000 \pm 0.000$ \\
& length 8 & $0.999 \pm 0.002$ & $-$ & $1.000 \pm 0.000$ \\
& length 12 & $0.965 \pm 0.025$ & $-$ & $1.000 \pm 0.000$ \\
& length 16 & $0.880 \pm 0.035$ & $-$ & $1.000 \pm 0.000$ \\
& length 20 & $0.759 \pm 0.043$ & $-$ & $1.000 \pm 0.000$ \\
& length 24 & $0.622 \pm 0.038$ & $-$ & $1.000 \pm 0.000$ \\
& length 28 & $0.484 \pm 0.043$ & $-$ & $0.999 \pm 0.000$ \\
& length 32 & $0.375 \pm 0.030$ & $-$ & $0.999 \pm 0.000$ \\
\midrule
\multirow{9}{*}{LPS}
& length 4 & $1.000 \pm 0.000$ & $1.000 \pm 0.000$ & $-$ \\
& length 7 & $1.000 \pm 0.000$ & $1.000 \pm 0.000$ & $-$ \\
& length 8 & $1.000 \pm 0.000$ & $-$ & $1.000 \pm 0.000$ \\
& length 16 & $0.999 \pm 0.001$ & $-$ & $1.000 \pm 0.000$ \\
& length 24 & $0.950 \pm 0.019$ & $-$ & $1.000 \pm 0.000$ \\
& length 32 & $0.788 \pm 0.019$ & $-$ & $1.000 \pm 0.000$ \\
& length 40 & $0.608 \pm 0.023$ & $-$ & $1.000 \pm 0.000$ \\
& length 48 & $0.477 \pm 0.030$ & $-$ & $0.999 \pm 0.001$ \\
& length 56 & $0.365 \pm 0.029$ & $-$ & $0.998 \pm 0.000$ \\
\midrule
\multirow{6}{*}{0-1 Knapsack}
& 2 items & $1.000 \pm 0.000$ & $1.000 \pm 0.000$ & $1.000 \pm 0.000$ \\
& 4 items & $0.966 \pm 0.006$ & $1.000 \pm 0.000$ & $1.000 \pm 0.000$ \\
& 6 items & $0.849 \pm 0.007$ & $-$ & $1.000 \pm 0.000$ \\
& 8 items & $0.640 \pm 0.242$ & $-$ & $1.000 \pm 0.000$ \\
& 10 items & $0.481 \pm 0.279$ & $-$ & $1.000 \pm 0.000$ \\
& 12 items & $0.435 \pm 0.252$ & $-$ & $0.988 \pm 0.029$ \\
\midrule
\multirow{6}{*}{MCM}
& 2 matrices & $0.973 \pm 0.009$ & $1.000 \pm 0.000$ & $1.000 \pm 0.000$ \\
& 4 matrices & $0.177 \pm 0.069$ & $-$ & $1.000 \pm 0.000$ \\
& 6 matrices & $0.088 \pm 0.029$ & $-$ & $1.000 \pm 0.000$ \\
& 8 matrices & $0.033 \pm 0.025$ & $-$ & $1.000 \pm 0.000$ \\
& 10 matrices & $0.051 \pm 0.032$ & $-$ & $0.998 \pm 0.001$ \\
& 12 matrices & $0.026 \pm 0.011$ & $-$ & $0.996 \pm 0.002$ \\
\bottomrule
\end{tabular}
\caption{The exact values of the Transformer experiments in Figure \ref{fig:exp:transformer} (algorithmic problems).}
\label{tab:val:transformer-alg}
\end{table}

\begin{table}
\centering
\small
\begin{tabular}{l|l|rrr}
\toprule
Problem & Difficulty & WT & CoT & RoT \\
\midrule
\multirow{8}{*}{Addition}
& 2-digit & $1.000 \pm 0.000$ & $1.000 \pm 0.000$ & $1.000 \pm 0.000$ \\
& 4-digit & $0.642 \pm 0.305$ & $1.000 \pm 0.001$ & $1.000 \pm 0.000$ \\
& 6-digit & $0.005 \pm 0.008$ & $0.997 \pm 0.005$ & $0.999 \pm 0.000$ \\
& 8-digit & $0.000 \pm 0.000$ & $0.905 \pm 0.155$ & $0.999 \pm 0.001$ \\
& 10-digit & $0.000 \pm 0.000$ & $0.795 \pm 0.341$ & $0.986 \pm 0.024$ \\
& 12-digit & $0.000 \pm 0.000$ & $-$ & $0.871 \pm 0.275$ \\
& 14-digit & $0.000 \pm 0.000$ & $-$ & $0.358 \pm 0.430$ \\
& 16-digit & $0.000 \pm 0.000$ & $-$ & $0.120 \pm 0.202$ \\
\midrule
\multirow{8}{*}{Subtraction}
& 2-digit & $1.000 \pm 0.000$ & $1.000 \pm 0.000$ & $1.000 \pm 0.000$ \\
& 4-digit & $0.776 \pm 0.179$ & $1.000 \pm 0.000$ & $1.000 \pm 0.000$ \\
& 6-digit & $0.006 \pm 0.001$ & $1.000 \pm 0.000$ & $1.000 \pm 0.000$ \\
& 8-digit & $0.000 \pm 0.000$ & $0.896 \pm 0.252$ & $0.994 \pm 0.016$ \\
& 10-digit & $0.000 \pm 0.000$ & $0.443 \pm 0.377$ & $0.908 \pm 0.236$ \\
& 12-digit & $0.000 \pm 0.000$ & $-$ & $0.507 \pm 0.398$ \\
& 14-digit & $0.000 \pm 0.000$ & $-$ & $0.295 \pm 0.406$ \\
& 16-digit & $0.000 \pm 0.000$ & $-$ & $0.101 \pm 0.137$ \\
\midrule
\multirow{7}{*}{Multiplication}
& 2-digit & $1.000 \pm 0.000$ & $1.000 \pm 0.000$ & $1.000 \pm 0.000$ \\
& 3-digit & $0.855 \pm 0.044$ & $-$ & $1.000 \pm 0.000$ \\
& 4-digit & $0.636 \pm 0.061$ & $-$ & $1.000 \pm 0.000$ \\
& 5-digit & $0.338 \pm 0.063$ & $-$ & $1.000 \pm 0.000$ \\
& 6-digit & $0.270 \pm 0.030$ & $-$ & $0.987 \pm 0.008$ \\
& 7-digit & $0.162 \pm 0.025$ & $-$ & $0.896 \pm 0.105$ \\
& 8-digit & $0.138 \pm 0.025$ & $-$ & $0.670 \pm 0.208$ \\
\midrule
\multirow{8}{*}{Division}
& 1-digit & $1.000 \pm 0.000$ & $1.000 \pm 0.000$ & $1.000 \pm 0.000$ \\
& 2-digit & $1.000 \pm 0.000$ & $-$ & $1.000 \pm 0.000$ \\
& 3-digit & $1.000 \pm 0.001$ & $-$ & $1.000 \pm 0.000$ \\
& 4-digit & $0.891 \pm 0.072$ & $-$ & $1.000 \pm 0.000$ \\
& 5-digit & $0.516 \pm 0.077$ & $-$ & $0.998 \pm 0.004$ \\
& 6-digit & $0.308 \pm 0.069$ & $-$ & $0.996 \pm 0.007$ \\
& 7-digit & $0.192 \pm 0.028$ & $-$ & $0.958 \pm 0.036$ \\
& 8-digit & $0.115 \pm 0.015$ & $-$ & $0.914 \pm 0.090$ \\
\bottomrule
\end{tabular}
\caption{The exact values of the LSTM experiments in Figure \ref{fig:exp:lstm}.}
\label{tab:val:lstm}
\end{table}

\end{document}